\documentclass[preprint,12pt]{elsarticle}

\usepackage{amssymb}
\usepackage{epstopdf}
\usepackage[margin=1in,showframe=false]{geometry}
\usepackage{graphicx}
\graphicspath{{./figs/}}
\usepackage{amsmath}
\usepackage{amsfonts}
\usepackage{mathtools}
\mathtoolsset{showonlyrefs}
\usepackage{bm}
\usepackage{xcolor}
\usepackage{hyperref}

\newcommand{\E}{\mathbb{E}}
\newcommand{\Var}{\mathrm{Var}}
\newcommand{\Cov}{\mathrm{Cov}}

\journal{arxiv}

\begin{document}

\begin{frontmatter}

%% Title, authors and addresses

%% use the tnoteref command within \title for footnotes;
%% use the tnotetext command for theassociated footnote;
%% use the fnref command within \author or \address for footnotes;
%% use the fntext command for theassociated footnote;
%% use the corref command within \author for corresponding author footnotes;
%% use the cortext command for theassociated footnote;
%% use the ead command for the email address,
%% and the form \ead[url] for the home page:
%% \title{Title\tnoteref{label1}}
%% \tnotetext[label1]{}
%% \author{Name\corref{cor1}\fnref{label2}}
%% \ead{email address}
%% \ead[url]{home page}
%% \fntext[label2]{}
%% \cortext[cor1]{}
%% \address{Address\fnref{label3}}
%% \fntext[label3]{}

\title{Gaussian Process Regression constrained by Boundary Value Problems}

\author[2]{M. Gulian}
\ead{mgulian@sandia.gov}
\author[1]{A. Frankel}
\ead{alfrank@sandia.gov}
\author[2]{L. Swiler}
\ead{lpswile@sandia.gov}
\address[1]{Computational Science and Analysis, Sandia National Laboratories, Livermore, CA, 94550}
\address[2]{Center for Computing Research, Sandia National Laboratories, Albuquerque, NM, 87123}

\begin{abstract}
We develop a framework for Gaussian processes regression constrained by boundary value problems. The framework may be applied to infer the solution of a well-posed boundary value problem with a known second-order differential operator and boundary conditions, but for which only scattered observations of the source term are available. Scattered observations of the solution may also be used in the regression. The framework combines co-kriging with the linear transformation of a Gaussian process together with the use of kernels given by spectral expansions in eigenfunctions of the boundary value problem. Thus, it benefits from a reduced-rank property of covariance matrices. We demonstrate that the resulting framework yields more accurate and stable solution inference as compared to physics-informed Gaussian process regression without boundary condition constraints.  
\end{abstract}

%%Graphical abstract
%\begin{graphicalabstract}
%\includegraphics{grabs}
%\end{graphicalabstract}

\iffalse
%%Research highlights
\begin{highlights}
\item Novel framework that combines boundary condition constraints via covariance kernels approximated by spectral expansions and PDE constraints via a co-kriging setup, allowing for Gaussian process regression constrained by boundary value problems 
\item First demonstration of Gaussian processes regression constrained by Neumann boundary conditions, as well as inference of the solution from scattered observations of the source term alone
\end{highlights}
\fi

\begin{keyword}
Scientific machine learning \sep constrained Gaussian process \sep physics-informed \sep boundary value problem \sep boundary condition
%% keywords here, in the form: keyword \sep keyword

%% PACS codes here, in the form: \PACS code \sep code

%% MSC codes here, in the form: \MSC code \sep code
%% or \MSC[2008] code \sep code (2000 is the default)

\end{keyword}

\end{frontmatter}

%% \linenumbers

%% main text

\section{Introduction}
Several physical processes are described by a well-posed boundary value problem (BVP) of the form
\begin{equation}\label{eq:baby_BVP}
\begin{split}
\left\{
\begin{aligned}
&L u(x) = f(x), \quad x \in \Omega, \\
&\mathcal{B}u(x) = g(x), \quad x \in \partial \Omega,
\end{aligned}
\right.
\end{split}
\end{equation}
where $L$ denotes a linear partial differential operator, $\Omega$ a domain with boundary $\partial \Omega$, and $\mathcal{B}$ a general mixed boundary operator. 
Prominent examples include the Poisson equation for electrostatics, advection-diffusion of a scalar (such as temperature or species concentration), time-harmonic wave propagation, or elastic deformation of materials. Under various conditions for $\Omega$, $L$ and $\mathcal{B}$, the resulting problem admits a unique solution for broad classes of source term $f$ and boundary term $g$. Therefore, when that data is fully known, deterministic solvers provide an ideal way to compute the solution $u$. However, while the operator $L$ and boundary condition are often known \emph{a priori}, in many cases the source term $f$ is only partially known from scattered observations in $\Omega$. In this case, the solution is no longer uniquely determined, and \emph{inference} must be used to estimate $u$ from the available data. Since observations of $f$ in $\Omega$ may be difficult to acquire and therefore sparse, the inference should make full use of the constraint represented by the BVP \eqref{eq:baby_BVP}, while providing an estimate of uncertainty associated with the prediction.

Gaussian process regression (GPR) is a widely used Bayesian technique for inference in scientific applications. Compared to other machine learning algorithms, GPR is especially suited for data that is limited or expensive, as it allows building a model that incorporates both prior information and observational data while naturally providing estimates for uncertainty quantification. Although deep learning can provide remarkable predictive capability in a wide array of tasks, in such applications the lack of a sufficiently large dataset can make it difficult to infer the parameters of deep neural networks. Obtaining uncertainty estimates in the prediction of a deep neural network requires an additional level of complexity. 

Physics-informed machine learning models which embed physical constraints are a highly active area of research \citep{raissi2018, lusch2018deep, ling2016machine,jones2018machine}.
Constraints for deep neural networks typically take the form of penalty terms in the loss functions to steer the model towards a more physically consistent one during training \citep{raissi2019physics,mao2020physics,cvpinns}. While simple to implement, it is then difficult to quantify the violation of the constraints when extrapolating. In contrast, a recent review of constrained GPR by \citet{swiler2020survey} revealed several approaches in the literature to enforce wide variety of constraints on a Gaussian processes, such as bound, shape, and linear differential equation constraints. The strategies ranged from methods that enforce the constraints at finite sets of ``virtual points,'' to more structural methods that enforce constraints ``globally'' over an entire input space by virtue of transforms or specific covariance kernels. 

The work of \citet{raissi2017} studied linear differential equation constraints of the form $Lu(x) = f(x)$ for GPR of a function $u(x)$ through a ``co-kriging'' setup when scattered observations of $u(x)$ and the forcing term $f(x)$ were available, extending the approach of \citet{graepel2003solving} which considered the case of observations of $f$ only.  
\citet{solin2019know} demonstrated that zero Dirichlet boundary values can be enforced in GPR by using a covariance kernel expanded in the Dirichlet eigenfunctions of the Laplacian. Rather than merely adding scattered observations of the boundary values, they obtained a noiseless, global enforcement of the boundary condition over $\partial\Omega$, while allowing observations of $u(x)$ within $\Omega$ to be noisy. The reduction of the regression to a finite eigenbasis also represented a compression of the dataset and led to speed-up due to the need to invert a much smaller matrix. 

We combine the covariance kernels of \citet{solin2019know} for boundary conditions with the differential equation constraints of \citet{raissi2017} within $\Omega$ to obtain a GPR model constrained by full well-posed BVP of the form \eqref{eq:baby_BVP}.  Coupling these two approaches together is a unique contribution. While \citet{graepel2003solving} considered boundary conditions together with differential equation constraints, the approach involved performing regression for a factorized representation of the solution which was constructed in special domains and for Dirichlet conditions; a straightforward construction for general domains was not provided, nor was co-kriging considered. Other related work include those of \citet{owhadi2015bayesian}, who considered Bayesian numerical homogenization, and \citet{albert2020gaussian}, who utilized covariance kernels in the form of a Mercer expansion to enforce PDE constraints. In addition to our unique framework, we go beyond the examples of \citet{solin2019know} by considering general mixed boundary conditions, such as Dirichlet conditions in certain regions of $\partial\Omega$ and Neumann conditions in other regions. Moreover, by combining the approaches in the manner presented here, we also achieve a speed-up and regularization of the physics-informed GPR approach of \citet{raissi2017} that honors the boundary conditions exactly. 

We review the standard approach and basic steps of GPR in Section \ref{sec:gpr}, noting where computational bottlenecks exist. Section \ref{sec:method} specifies the types of BVPs that we consider and gives an overview of our framework. There, we first review the differential equation constraints of \citet{raissi2017} and boundary condition constraints of \citet{solin2019know}. We see that because these constraints affect different steps of GPR, they can be combined in a straightforward way. Section \ref{sec:examples} demonstrates the methodology on simple example problems, including comparisons with PDE-constrained GPR without boundary condition constraints. The advantages of the framework are then summarized in Section \ref{sec:conclusion}.  

To provide an overview of the types of GPs we consider, we list the four approaches discussed and compared in this paper for inferring the solution $u$ to a boundary value problem of the form \eqref{eq:baby_BVP}. 
\begin{itemize}
\item \textbf{Unconstrained GPR.}  This is typical GPR as presented in Section \ref{sec:gpr}, with no constraints implemented. We use a squared exponential kernel for the examples presented in Section \ref{sec:examples}.  Only observations of the solution $u$ are used. 
\item \textbf{Boundary Constrained GPR (BC-GPR)}.  This formulation, presented in Section \ref{sec:bc_constrained_gpr}, uses a special covariance kernel to satisfy a known boundary condition.  Specifically, the covariance is represented by a spectral expansion involving the eigenfunctions of the solution to the BVP. Only observations of the forcing term $f$ are used.
\item \textbf{Linear Partial Differential Equation constrained GPR (PDE-GPR)}.  This formulation, presented in Section \ref{sec:pde_constrained_gpr}, utilizes a known linear operator in a PDE relating the solution $u$ to a forcing term $f$. Co-kriging is used to capture the relationship the solution $u$ and forcing term $f$; scattered observations of $u$ and $f$ are utilized. 
\item \textbf{Boundary Value Problem-Constrained GPR (BVP-GPR)}. This is the approach we present in Section \ref{sec:bvp_constrained_gpr}, which combines aspects of BC-GPR and PDE-GPR to treat both boundary condition and linear PDE constraints.  The covariance kernels are approximated by spectral expansion involving eigenfunctions and the co-kriging approach is used.  Observations of the solution $u$, forcing $f$, or both may be used.    
\end{itemize}

\section{Gaussian Process Regression}\label{sec:gpr}

This section reviews the basics of GPR. More extensive reviews can be found in \citet{rasmussen} and \citet{murphy2012machine}. Seminal work discussing the use of GPs as surrogate models for computational science and engineering applications include \citet{sacks} and \citet{santner}. 
In GPR, we assume that an underlying function of interest $u(x)$ is modeled by a Gaussian process with a given mean function $m(x)$ and covariance function between any two points $x$ and $x'$ given by $K(x,x') = \Cov(u(x),u(x'))$:
\begin{equation}
u\sim \mathcal{G}\mathcal{P}(m,K).
\end{equation}
That is, the vector of values $u(X)$ over a finite collection of locations $X$ has a multivariate normal density
\begin{equation}\label{eq:gp_prior_def}
u(X) \sim \mathcal{N}(m(X),K(X,X)),
\end{equation}
where $m(X)$ is a vector of mean values of $u$ and $K(X,X)$ is the covariance matrix between the values. One common choice of the covariance function is the squared-exponential kernel given by
\begin{equation}\label{eq:squared_exponential}
K(x,x') = s^2 \exp\left(-\frac{|x-x'|^2}{2 \ell^2}\right) 
\end{equation}
where $s^2$ and $ \ell^2$ are magnitude and length-scale parameters that control the behavior of the covariance function, i.e., the hyperparameters. 

Without loss of generality, we assume the mean function $m$ is zero as the Gaussian process posterior is known to satisfy statistical consistency.
Then we can write the density function for a GP prior distribution \eqref{eq:gp_prior_def} over $N$ points as
\begin{equation}\label{eq:prior}
p(u|X) = (2\pi)^{-N/2}|K|^{-1/2}\exp\left(-\frac{1}{2}u^\top K^{-1}u\right), \quad K = K(X,X).
\end{equation}
We assume that data or observations $y$ at the $X$ locations are contaminated by independently and identically distributed Gaussian noise with variance $\sigma^2$, giving a likelihood function
\begin{equation}\label{eq:likelihood}
p(y|u,X) = \prod_{i=1}^N \frac{1}{\sqrt{2\pi \sigma^2}}\exp\left(-\frac{(y_i-u_i(X_i))^2}{2\sigma^2}\right).
\end{equation}
Gaussian process regression proceeds by invoking Bayes' rule to compute the posterior distribution of $f$ as
\begin{equation}
p(u|y,X) = \frac{p(y|u,X)p(u|X)}{p(y|X)},
\end{equation}
with log-marginal-likelihood
\begin{align}\label{eq:lml}
\begin{split}
\log p(y|X) &= \int p(y|u,X) p(u|X) du \\
&= 
\begin{multlined}[t]
-\frac{1}{2}y^\top (K(X,X)+\sigma^2 I_N)^{-1}y -\frac{1}{2}\log|K(X,X)+\sigma^2 I_N| - \frac{N}{2}\log 2\pi,
\end{multlined}
\end{split}
\end{align}
using the prior \eqref{eq:prior} and the Gaussian likelihood \eqref{eq:likelihood}.
Here, $I_N$ denotes the identity matrix of size $N\times N$. The predictive distribution for $u^*=u(x^*)$ at a new point $x^*$ can then be shown \citep{rasmussen,murphy2012machine} to be a Gaussian with mean
\begin{equation}\label{eq:posterior_mean}
\E[u^*] = K(x^*,X)(K(X,X)+\sigma^2 I_N)^{-1}y
\end{equation}
and variance
\begin{equation}\label{eq:posterior_variance}
\Var[u^*] = K(x^*,x^*) - K(x^*,X)(K(X,X)+\sigma^2 I_N)^{-1}K(X,x^*).
\end{equation}

Given a set of hyperparameters for the covariance function, inference through the above equations is straightforward albeit costly. Standard approaches leverage the Cholesky decomposition of the matrix $K(X,X)+\sigma^2 I_N$ for numerical stability, which has cost $O(N^3)$, and can be used to compute the matrix inverse and log-determinant in the marginal likelihood.
In practice, one rarely knows the hyperparameters for the covariance function. The most common way of handling this is to use maximum likelihood optimization of the log-marginal-likelihood with respect to the covariance hyperparameters. To enable quasi-Newton methods, an analytical gradient calculation of the log-marginal-likelihood \eqref{eq:lml} with respect to arbitrary hyperparameters $\theta_k$ and the noise variance $\sigma^2$ is given by
\begin{align}
\label{eq:MLder}
\frac{\partial \log p(y|X)}{\partial \theta_k} &= \frac{1}{2}y^\top \tilde{K}^{-1}\frac{\partial K}{\partial \theta_k}\tilde{K}^{-1}y - \frac{1}{2}\text{Tr}\left(\tilde{K}^{-1}\frac{\partial K}{\partial \theta_k}\right) \\
\label{eq:MLder2}
\frac{\partial \log p(y|X)}{\partial \sigma^2} &= \frac{1}{2}y^\top \tilde{K}^{-1}\tilde{K}^{-1}y -\frac{1}{2}\text{Tr}(\tilde{K}^{-1})
\end{align}
where $\tilde{K}=K+\sigma^2 I_N$, and $\text{Tr}$ is the trace operator. Some shortcuts are available, e.g. by performing multiple matrix-vector multiplications in advance and successively multiplying matrices on to the resulting vectors, and by only computing the diagonal vector dot products in the trace terms. However, since the objective function \eqref{eq:lml} is not convex, multiple restarts from different initial points may be needed to guarantee an acceptable optimum has been found, and the resulting hyperparameter optimization cost may be very expensive.

\section{Boundary Value Problem Constraints}\label{sec:method}

\subsection{Problem Statement}\label{sec:problem}

We consider GPR of a function $u(x)$ that is known to satisfy a boundary value problem with mixed boundary conditions
\begin{equation}
\begin{split}\label{eq:BVP}
\left\{
\begin{aligned}
&L u(x) = f(x), \quad x \in \Omega, \\
&a_i u(x) + b_i \nabla u(x) \cdot \hat{n}(x) = 0, \quad x \in \Gamma_i, \quad i=1, ..., n.
\end{aligned}
\right.
\end{split}
\end{equation}
Here, $L$ denotes a second-order linear differential operator of the form 
\begin{equation}\label{eq:elliptic_operator}
L u (x) = \sum_{i,j} a_{ij}(x) \frac{\partial^2 u}{\partial x_i \partial x_j}(x) + 
\sum_{i} b_{i}(x) \frac{\partial u}{\partial x_i}(x) +
c(x) u(x), 
\end{equation}
$a_i$ and $b_i$ are constants dependent on $i$, and 
\begin{equation}
\partial\Omega = \Gamma_1 \cup \Gamma_2 \cdots \cup \Gamma_n, \text{ for disjoint } \{\Gamma_i\}_{i=1}^n.
\end{equation}
We assume $L$ and the boundary conditions are known everywhere, and that scattered observations of $u$ and/or $f$ are available in the interior of the domain $\Omega$. Note that $a_i = 1, b_i = 0$ would yield Dirichlet conditions over the portion $\Gamma_i$ of the boundary $\partial \Omega$, while $a_1 = 0, b_i = 1$ would yield Neumann conditions; see Section \ref{sec:2d} for an example.
There is no loss of generality in considering zero boundary value in \eqref{eq:BVP}, since the solution to an inhomogeneous
boundary value problem
\begin{equation}
\begin{split}\label{eq:BVP_inhomogeneous}
\left\{
\begin{aligned}
&L u(x) = f(x), \quad x \in \Omega, \\
&a_i u(x) + b_i \nabla u(x) \cdot \hat{n}(x) = g(x), \quad x \in \Gamma_i, \quad i=1, ..., n,
\end{aligned}
\right.
\end{split}
\end{equation}
for $g \neq 0$ can be written as $u = u_1 + u_2$, where $u_1$ solves \eqref{eq:BVP} and $u_2$ solves the problem
\begin{equation}
\begin{split}\label{eq:BVP_no_source}
\left\{
\begin{aligned}
&L u(x) = 0, \quad x \in \Omega, \\
&a_i u(x) + b_i \nabla u(x) \cdot \hat{n}(x) = g(x), \quad x \in \Gamma_i, \quad i=1, ..., n.
\end{aligned}
\right.
\end{split}
\end{equation}
Thus, $u_2$ can be solved for exactly, and inference is only required for $u_1$ using the scattered observations of $f$.

\subsection{Linear Differential Equation Constraints}\label{sec:pde_constrained_gpr}
In this section, we provide an overview of how Gaussian processes may be constrained to satisfy the differential equation constraint in \eqref{eq:BVP} using scattered observations of $f$ and $u$ with the approach of \citet{raissi2017}.  We note that the observations of $f$ and $u$ may occur at the same or different $x$ points.
If $u(x)$ is a GP with mean function $m(x)$ and covariance kernel
$k(x,x)$,
\begin{equation}
\label{eq:gp_for_u}
u \sim \mathcal{G}\mathcal{P}(m(x),k(x,x')),
\end{equation}
and if $m(\cdot)$ and $k(\cdot, x')$ belong to the domain of the operator $L$ given by \eqref{eq:elliptic_operator}, then
$L_x L_{x'} k(x,x')$ defines a valid covariance kernel for a
GP with mean function $L_x m(x)$.
This Gaussian process is denoted $Lu$:
\begin{equation}
\label{eq:gp_for_Lu}
L u \sim \mathcal{G}\mathcal{P}(L_x m(x), L_x L_{x'} k(x,x')).
\end{equation}
If scattered measurements
$y_f$ %$Y_f = \{\mathbf{y}_f\}$
on the source term $f$ in \eqref{eq:BVP} are available at $N_f$ domain points $X_f$,
% = \{\mathbf{x}_f\}$
then this can be used to train and obtain predictions for $Lu$ from the GP \eqref{eq:gp_for_Lu}  in the standard way. If, in addition, measurements $y_u$
%$Y_u = \{\mathbf{y}_u\}$
of $u$ are available at $N_u$ domain points $X_u$
% = \{\mathbf{x}_u\}$,
a GP co-kriging procedure can be used. In this setting physics knowledge encoded in the differential equation in \eqref{eq:BVP} enters via the data $(X_f,y_f)$ and can be used to improve prediction accuracy and reduce the variance of the GP for $u$.

The co-kriging procedure requires forming the joint Gaussian process $[u(x_1);f(x_2)]$.
The covariance matrix of the resulting GP is a four block matrix assembled from
the covariance matrix of the GP \eqref{eq:gp_for_u} for the solution $u$, the covariance of the GP \eqref{eq:gp_for_Lu} for the forcing function, and the cross terms.
Given the covariance kernel $k(x,x')$ for $u$, the covariance kernel of this joint GP
is
\begin{equation}\label{eq:joint_covariance}
k
\left(
\begin{bmatrix}
x_1 \\
x_2
\end{bmatrix}
,
\begin{bmatrix}
x'_1 \\
x'_2
\end{bmatrix}
\right)
=
\begin{bmatrix}
\phantom{L_x} k(x_1,{x}'_1) &
\phantom{L_x L_{x'}} L_{x'}k({x}_1,{x}'_2)\\
L_x  k ({x}_2,{x}'_1)
& L_x L_{x'} k({x}_2,{x}'_2)
\end{bmatrix}
=
\begin{bmatrix}
K_{11} &
K_{12} \\
K_{21} &
K_{22}
\end{bmatrix}.
\end{equation}

The covariance between $u({x})$ and $f({x'})$ is given by $L_{x'}  k({x},{x}')$ in the upper right block of the kernel; see \citet{raissi2017}. Similarly the covariance between $u({x}')$ and $f(x)$ is represented by the bottom left block $L_x k({x},{x}')$ of the kernel.
The joint Gaussian process for $[u; f]$ is then
\begin{equation}\label{eq:joint_GP_uf}
\begin{bmatrix}
u(x_1) \\
f(x_2)
\end{bmatrix}
\sim \mathcal{G}\mathcal{P}\left(
\begin{bmatrix}
\phantom{\mathcal{L}}
m(x_1) \\
L m(x_2)
\end{bmatrix},
\begin{bmatrix}
K_{11}(x_1,x_1) &
K_{12}(x_1,x_2) \\
K_{21}(x_2,x_1) &
K_{22}(x_2,x_2)
\end{bmatrix}
\right),
\end{equation}
where $K_{12}(x_1,x_2) = \left[K_{21}(x_2,x_1)\right]^\top$. Given $N_u$ observations ($X_u,y_u$) of $u$ and $N_f$ observations ($X_f,y_f$) of $f$, formulas \eqref{eq:posterior_mean} and \eqref{eq:posterior_variance} then can be used for posterior prediction for 
$[u(x^*); f(x^*)]$. Alternatively \citep{raissi2017}, prediction for $u^* = u(x^*)$ separately can be written as 
\begin{align}
\E[u^*] &= Q_u (K(X,X)+\sigma^2 I_{N_u})^{-1} y \\
\Var[u^*] &= K_{11}(x^*,x^*) - Q_u (K(X,X)+\sigma^2 I_{N_u})^{-1}Q_u^\top,
\end{align}
where $X = [X_u; X_f]$, $K(X,X)$ is the covariance in \eqref{eq:joint_GP_uf}, $y = [y_u; y_f]$, and 
\begin{equation}
Q_u = [ K_{11}(x^*,X_u) \ K_{12}(x^*,X_f) ].
\end{equation}
Note that this can be used to infer $u^*$ when no observations are available for $u$, but observations are available for $f$; in this case, $y = y_f$, $K(X,X) = K_{22}(X_f,X_f)$, and
$Q_u = K_{12}(x^*,X_f)$. However, as discussed in Section \ref{sec:1d}, this performs poorly with standard covariance kernels, as opposed to the covariance kernels discussed in Section \ref{sec:bc_constrained_gpr} that are informed by known boundary conditions.

The hyperparameters governing equation \eqref{eq:joint_GP_uf} can be trained by minimizing the negative log-marginal-likelihood as outlined
in \eqref{eq:MLder} and \eqref{eq:MLder2}. Without loss of generality, we will assume that $m(x) = 0$ on the grounds that provided enough data, the GPR will converge to the correct response regardless of the provided mean function, although a judiciously chosen mean function can improve extrapolation or convergence.\\

\subsection{Eigenfunction Expansion Kernel Functions for Boundary Conditions}\label{sec:bc_constrained_gpr}
To perform Gaussian process regression, we suppose that the behavior of the function of interest is described by some covariance function $k(x,x')$.
The posterior mean prediction \eqref{eq:posterior_mean} in GPR for a function $u$ at a point $x$, given data $(X,y) = \{(x_i, y_i)\}_{i=1}^N$, can be written as
\begin{equation}\label{eq:representer}
\E[u(x)] = \sum_{i=1}^N c_i k(x,x_i),
\end{equation}
for coefficients $c_i \in \mathbb{R}^d$ that depend on the covariance kernel function $k$, the hyperparameters, and the data $(X,y)$. This property is an instance of the representer theorem for reproducing kernel Hilbert spaces \citep{kimeldorf1971some,berlinet2011reproducing}. It implies that, if for all $x'$, the function $k(x, x')$ satisfies satisfies the homogeneous boundary condition of the BVP \eqref{eq:BVP} for all $x \in \partial\Omega$, then so will the posterior mean prediction $\E[u(x)]$. This suggests the use of covariance kernel functions that satisfy the boundary conditions, as opposed to standard choices such as the squared-exponential kernel\eqref{eq:squared_exponential}, for GPR with boundary condition constraints. 

The spectral theory of elliptic operators \citep{davies95, edmunds2018} provides a framework for general construction of such kernels. This theory provides a variety of conditions under which the solution of the BVP \eqref{eq:BVP} can be expanded in orthonormal eigenfunctions defined by
\begin{equation}\label{eq:eigenvalue_problem}
\left\{
\begin{aligned}
&L \phi_n(x) = \lambda_n \phi_n(x), \quad x \in \Omega, \\
&a_i \phi_n(x) + b_i \nabla \phi_n(x) \cdot \hat{n}(x) = 0,  \quad x \in \Gamma_i, \quad i=1, ..., n,
\end{aligned}
\right.
\end{equation}
for some eigenvalues $\lambda_n$ and orthonormal eigenfunctions $\phi_n$. 
The specific conditions vary depending on the type of boundary condition, and involve the reglarity of the domain $\Omega$ and coefficients of the operator $L$.
Any convergent expansion in $\phi_n(x) \phi_{n'}(x')$ will then satisfy the boundary conditions. \citet{solin2019know} and \citet{solin2019hilbert} proposed that the covariance function be given by the specific expansion
\begin{equation}\label{eq:expansion_kernel}
k(x,x') = \sum_{n=1}^M S\left(\sqrt{\lambda_n}\right) \phi_n(x)\phi_n(x'),
\end{equation}
where $S\left(\sqrt{\lambda_n}\right)$ is the spectral power density (Fourier transform) of an ``original'' covariance function of interest. For example, for the squared-exponential covariance kernel \eqref{eq:squared_exponential}, the function is given by
\begin{equation}\label{eq:spectral_density}
S(\omega) = s^2 (2\pi \ell^2)^{d/2} \exp\left(-\frac{1}{2} \ell^2 \omega^2\right).
\end{equation}
\citet{solin2019hilbert} considered such expansions \eqref{eq:expansion_kernel}, \eqref{eq:spectral_density} using the Dirichlet spectrum of the Laplacian for reduced-rank approximation in unconstrained GPR, deriving that such expansions approximate the original covariance function as the boundaries approach infinity. 

\begin{figure}
\centering
\includegraphics[width=0.8\linewidth]{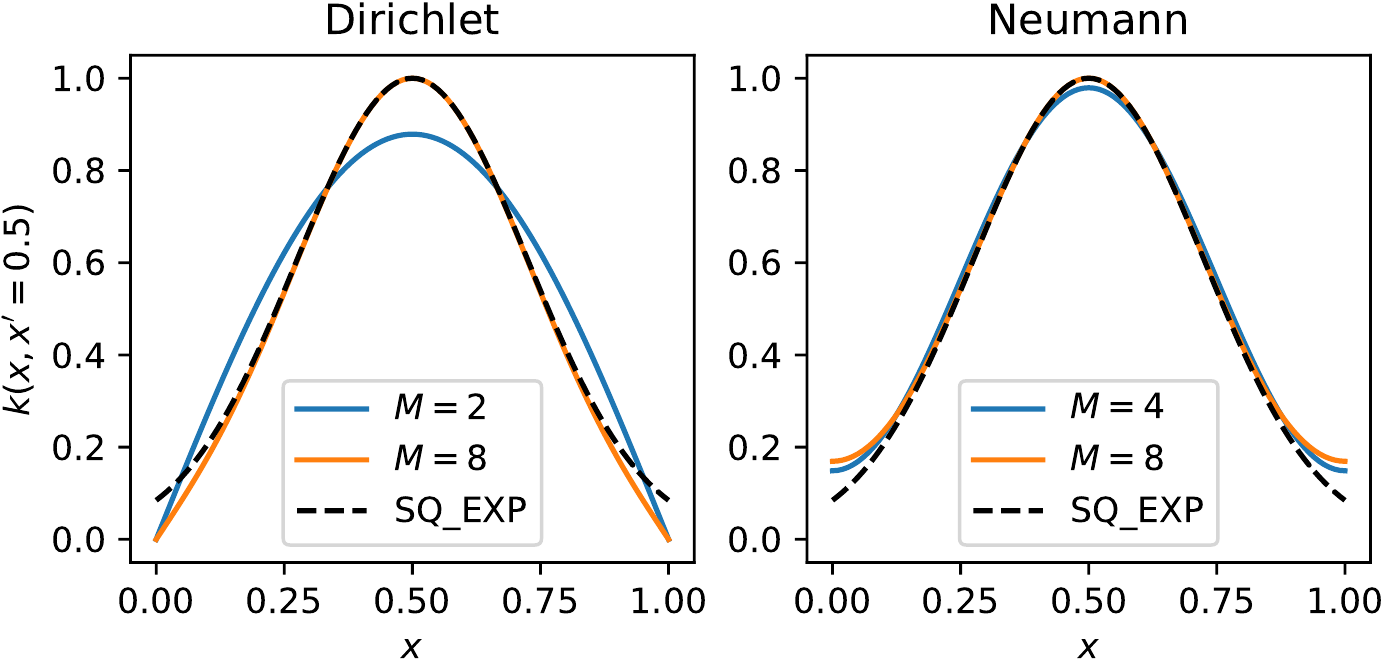}
\caption{Comparison of the squared-exponential kernel $k(x,x' = 0.5)$ with the corresponding spectral expansion kernel \eqref{eq:expansion_kernel} at $x' = 0.5$ for $x \in \Omega = (0,1)$, defined using homogeneous Dirichlet (\emph{left}) and Neumann (\emph{right}) spectrum for different $M$. The squared-exponential kernel satisfies neither zero Dirichlet nor zero Neumann boundary conditions. For $M=8$, the expansion kernels approximate the squared-exponential kernel in the interior of $[0,1]$, while diverging from it near the boundaries to satisfy the boundary conditions.}
\label{fig:kernel_approximation}
\end{figure}

For fixed domain $\Omega$, we note that \eqref{eq:expansion_kernel} is a convergent series if $S(\omega)$ exhibits sufficiently rapid decay, although the series cannot be expected to converge to the original kernel on $\overline{\Omega} = \Omega \cup \partial \Omega$ unless the original kernel itself satisfied the boundary conditions; see Figure \ref{fig:kernel_approximation}.  
\citet{solin2019know} first exploited the boundary values of such kernels to perform constrained GPR for Dirichlet problem for the Laplacian. In this context, the kernel \eqref{eq:expansion_kernel} can be thought of as a projection of the original kernel onto an appropriate space of functions satisfying the boundary conditions. As illustrated in Figure \ref{fig:kernel_approximation}, this yields a non-stationary covariance kernel that encodes the desired boundary conditions over all of $\partial\Omega$ while approximating the behavior of the original kernel in the interior $\Omega$. In practice, the series is truncated at some finite order $M$, where increasing order corresponds to increasing fidelity to the original or ``parent'' covariance function away from the boundary $\partial\Omega$. We have not found a rigorous statement and proof of this property in the literature, but the framework of \citet{solin2019know} that we review and combine with linear PDE constraints below is valid for any expansion of the form \eqref{eq:expansion_kernel}. It is beyond the scope of the this work to explore the theoretical properties of such kernels, and we simply use \eqref{eq:expansion_kernel} and \eqref{eq:spectral_density} throughout the rest of the paper.

\subsection{Combining Boundary Value and Linear PDE Constraints}\label{sec:bvp_constrained_gpr}

We return now to Gaussian process regression in the context of the spectral expansion method. Suppose we observe $N$ function value observations of $u(x_i)$ at different point locations $x_i \in \Omega, i=1, 2, ..., N$. Per the review of GPR in Section \ref{sec:gpr}, using a covariance kernel of the form \eqref{eq:expansion_kernel}, the covariance matrix augmented with the Gaussian likelihood (white noise) is given by
\begin{equation}\label{eq:expansion_matrix}
\tilde{K} = K+\sigma^2 I_N = \Phi \Lambda \Phi^\top + \sigma^2 I_N,
\end{equation}
where $\Phi$ is the $N \times M$ matrix of eigenfunctions at the point locations,
\begin{equation}
[\Phi]_{i,j} = \phi_j(x_i), \quad 1 \le i \le N, \quad 1 \le j \le M,
\end{equation}
and $\Lambda$ is the $M \times M$ diagonal matrix of the spectral power density evaluated at the eigenvalues $\lambda_j$ corresponding to the $\phi_j$,
\begin{equation}\label{eq:lambda_matrix}
\Lambda = \text{diag}\left(S\left(\sqrt{[\lambda_1 \ \lambda_2 \ ...  \ \lambda_M ]}\right)\right).
\end{equation}

As pointed out by \citet{solin2019hilbert}, if $M \ll N$, then not only does the spectral expansion approach expressed by \eqref{eq:expansion_matrix} satisfy the boundary value problem \eqref{eq:BVP}, the linear algebraic operations for the posterior prediction and maximum likelihood estimation of the hyperparameters may be rearranged to admit a more efficient implementation by use of the Woodbury matrix inversion lemma \citep{higham2002}. The inverse of the $N \times N$ covariance matrix \eqref{eq:expansion_matrix} can be calculated as
\begin{equation}\label{eq:nice_inverse}
\tilde{K}^{-1} = \frac{1}{\sigma^2}(I_N-\Phi Z^{-1}\Phi^\top),
\end{equation}
where we have defined the $M \times M$ matrix
\begin{equation}\label{eq:Z_def}
Z = \sigma^2 \Lambda^{-1} + \Phi^\top \Phi.
\end{equation}
Then, given a new prediction point location $x^*$ and defining the vector
\begin{equation}
\phi_* = [\phi_1(x^*) \ \phi_2(x^*) \ ... \ \phi_M(x^*)]^\top,
\end{equation}
the mean and variance of the GP posterior prediction for $f^* = f(x^*)$, formulas \eqref{eq:posterior_mean} and \eqref{eq:posterior_variance}, respectively, can be written as
\begin{align}\label{eq:reduced_posterior}
\E(f^*) &= \phi_*^\top Z^{-1}\Phi^\top y, \\
\Var(f^*) &= \sigma^2 \phi_*^\top Z^{-1}\phi_*.
\end{align}
The negative of the log marginal likelihood \eqref{eq:lml} can be written as 
\begin{equation}
-\log p(y|X) = \frac{1}{2}\log|\tilde{K}| + \frac{1}{2}y^\top \tilde{K}^{-1}y + \frac{N}{2}\log 2\pi
\end{equation}
where $\log|\tilde{K}|$ and its gradient are given by
\begin{align}\label{eq:log_term}
\log |\tilde{K}|
&= 
(N-M)\log \sigma^2 + \log |Z| + \sum_{j=1}^M \log \Lambda_{jj},
\\
\frac{\partial}{\partial \theta_k} \log|\tilde{K}|
&=
-\sigma^2 \text{Tr}\left(Z^{-1}\Lambda^{-2}\frac{\partial \Lambda}{\partial \theta_k}\right) + \sum_{j=1}^M \frac{1}{\Lambda_{jj}}\frac{\partial \Lambda_{jj}}{\partial \theta_k},
\\
\frac{\partial}{\partial \sigma^2} \log|\tilde{K}| 
&=
\frac{N-M}{\sigma^2} + \text{Tr}(Z^{-1}\Lambda^{-1}),
\end{align}
and the quadratic term $y^\top \tilde{K}^{-1}y$ and its gradient are given by
\begin{align}
y^\top \tilde{K}^{-1} y 
&=
\frac{1}{\sigma^2} (y^\top y - y^\top \Phi Z^{-1} \Phi^\top y),
\\
\frac{\partial}{\partial \theta_k} y^\top \tilde{K}^{-1} y &= -y^\top \Phi Z^{-1} \left( \Lambda^{-2}\frac{\partial \Lambda}{\partial \theta_k}\right) Z^{-1}\Phi^\top y,
\\
\label{eq:quadratic_term}
\frac{\partial}{\partial \sigma^2} y^\top \tilde{K}^{-1} y &= -\frac{1}{\sigma^4}(y^\top y - y^\top\Phi Z^{-1}\Phi^\top y)+\frac{1}{\sigma^2}y^\top \Phi Z^{-1}\Lambda^{-1}Z^{-1}\Phi^\top y.
\end{align}
Here, we point out a significant computational advantage when using formulas \eqref{eq:nice_inverse}--\eqref{eq:quadratic_term} for GPR. 
Assembly of $Z$ throughout these equations amounts to a computational complexity of $\mathcal{O}(NM^2)$, while inversion amounts to $\mathcal{O}(M^3)$ rather than $\mathcal{O}(N^3)$,  representing significant savings if only a few basis functions are used to represent the covariance kernel compared to the number of observations. 
The inverse and log determinant of $Z$ can be computed efficiently with a Cholesky decomposition, and the trace terms can be evaluated with elementwise multiplication.

We now show how the above framework extends to the co-kriging setup of the linear PDE constaints outlined in Section \ref{sec:pde_constrained_gpr}. We assume that we are given observations of both the function $u$ and the forcing term $f$ at potentially disjoint locations $X_u$ and $X_f$, respectively. We also assume that a kernel function of the form \eqref{eq:expansion_kernel} is used in which the eigenfunctions and eigenvalues are consistent with the BVP defining the constraint; that is, $\{(\lambda_i, \phi_i)\}$ solve \eqref{eq:eigenvalue_problem} for the same operator $L$ appearing in \eqref{eq:BVP}. Formally, we compute the covariance between the solution $u$ and forcing term $f$ as
\begin{align}\label{eq:cross_block}
\begin{split}
\Cov(u(x),f(x')) &= \Cov(u(x),Lu(x')) \\
&= \sum_{j=1}^M S\left(\sqrt{\lambda_j}\right)\phi_j(x) L\phi_j(x') \\
&= \sum_{j=1}^M S\left(\sqrt{\lambda_j}\right) \lambda_j\phi_j(x)\phi_j(x')
\end{split}
\end{align}
and between the forcing term and itself as
\begin{align}\label{eq:last_block}
\begin{split}
\Cov(f(x),f(x')) &= \Cov(Lu(x),Lu(x')) \\
&= \sum_{j=1}^M S\left(\sqrt{\lambda_j}\right) \lambda_j^2 \phi_j(x)\phi_j(x').
\end{split}
\end{align}

The covariance matrix between the solution and forcing observations can therefore be constructed in a block-matrix form as
\begin{equation}\label{eq:joint_GP_uf_spectral}
\begin{bmatrix}
u({X}_u) \\
f({X}_f)
\end{bmatrix}
\sim \mathcal{G}\mathcal{P}\left(
\begin{bmatrix}
\phantom{\mathcal{L}}
m({X}_u) \\
L m({X}_f)
\end{bmatrix},
K_{\text{joint}}
\right),
\end{equation}
where
\begin{equation}\label{eq:four_block_expansion}
K_{\text{joint}} =
\begin{bmatrix}
\phantom{\lambda_j} \sum_{j=1}^M S(\sqrt{\lambda_j})\phi_j(X_u)\phi_j(X_u)^\top &
\sum_{j=1}^M S(\sqrt{\lambda_j}) \lambda_j \phi_j(X_u)\phi_j(X_f)^\top\\
\sum_{j=1}^M S(\sqrt{\lambda_j}) \lambda_j \phi_j(X_f) \phi_j(X_u)^\top&
\sum_{j=1}^M S(\sqrt{\lambda_j}) \lambda_j^2 \phi_j(X_f)\phi_j(X_f)^\top\\
\end{bmatrix}.
\end{equation}
Defining the $N_u \times M$ matrix $\Phi_u$ and the $N_f \times M$ matrix $\Phi_f$ as
\begin{align}
[\Phi_u]_{i,j} = &\phi_j(x_i), \quad 1 \le i \le N_u, \quad x_i \in X_u, \quad 1 \le j \le M, \\
[\Phi_f]_{i,j} = \lambda_i &\phi_j(x_i), \quad 1 \le i \le N_f, \quad x_i \in X_f, \quad 1 \le j \le M,
\end{align}
and the block matrix
\begin{equation}
\Phi_{\text{joint}} = 
\begin{bmatrix}
\Phi_u \\
\Phi_f
\end{bmatrix},
\end{equation}
the covariance matrix \eqref{eq:four_block_expansion} augmented by the Gaussian likelihood can be written as 
\begin{equation}\label{eq:augmented_four_block}
\tilde{K}_{\text{joint}} = K_{\text{joint}}+\sigma^2 I_{N_u + N_f} = \Phi_{\text{joint}} \Lambda \Phi_{\text{joint}}^\top +\sigma^2 I_{N_u + N_f}.
\end{equation}
The form of this kernel mimics that of \eqref{eq:expansion_matrix}. Defining $Z$ as in \eqref{eq:Z_def} with $\Phi_{\text{joint}}$ in place of $\Phi$ allows the entire reduced-rank framework expressed by equations \eqref{eq:reduced_posterior}--\eqref{eq:quadratic_term} to be utilized, with the matrix $\Phi_{\text{joint}}$ in place of $\Phi$ throughout. In Section \ref{sec:examples}, we will study using this method to infer the solution $u$ using only observations of $f$; as explained at the conclusion of Section \ref{sec:pde_constrained_gpr}, in this setting only some of the blocks of co-kriging covariance matrix $K_{\text{joint}}$ are present, but the reduced-rank framework above represents this case just as well.  

The above formulation assumes implicitly that the measurement noise strength in the solution measurements and source term measurements are identical. \citet{raissi2017} considered linear PDE constrained GPR in which this is not true. In the above formulation, this would result in $K_{\text{joint}}$ being augmented to
\begin{equation}
\tilde{K}_{\text{joint}} =
\Phi_{\text{joint}} \Lambda \Phi_{\text{joint}}^\top +
\begin{bmatrix}
\sigma_u^2 I_{N_u} & 0 \\ 
0 & \sigma_f^2 I_{N_f}
\end{bmatrix}
=
\Phi_{\text{joint}} \Lambda \Phi_{\text{joint}}^\top + D
\end{equation}
rather than \eqref{eq:augmented_four_block}. 
The inverse of a matrix of the form $\tilde{K} = \Phi \Lambda \Phi^\top + D$ for a general diagonal matrix $D$ is given by the Woodbury lemma as
\begin{equation}
\tilde{K}^{-1} = D^{-1} (I-\Phi(\Lambda^{-1}+\Phi^\top D\Phi)^{-1}\Phi^\top),
\end{equation}
which is slightly more complicated compared to \eqref{eq:nice_inverse}, in which $D = \sigma^2 I$.  
Although this would complicate the evaluation of the marginal likelihood and gradient and require modified derivations of \eqref{eq:reduced_posterior}--\eqref{eq:quadratic_term}, there are still computational savings in this formulation. For demonstration purposes, we will limit ourselves to the identical variance case.

Imposing linear differential constraints of the form $Lu = f$ leaves the reduced-rank formulation of \citet{solin2019know} intact because we have assumed consistency between the spectrum 
$\{(\lambda_j, \phi_j)\}$ defining the kernel \eqref{eq:expansion_kernel} and the operator $L$. This is not necessary to obtain a BVP constrained GP. For example, the Dirichlet eigenfunctions of the Laplacian $\Delta$ form a complete orthonormal basis of $H^1_0(\Omega)$ \citep{larsson2008partial}. Thus, rather than solving \eqref{eq:eigenvalue_problem} for the spectrum of $L$, it would suffice use the Dirichlet eigenvalues and eigenfunctions of the Laplacian to define a covariance kernel of the form \eqref{eq:expansion_kernel}, and then apply the co-kriging setup of Section \ref{sec:pde_constrained_gpr} to impose $Lu = f$. While this would remove the requirement of solving \eqref{eq:eigenvalue_problem} for given $L$, formulas \eqref{eq:cross_block} and \eqref{eq:last_block} would no longer be valid, so direct inference using the four-block covariance matrix in \eqref{eq:joint_GP_uf} will be required. In the examples below, we consider covariance kernels constructed from the spectrum of $L$ to make use of the reduced-rank framework.

\section{Examples}\label{sec:examples}

In this section, we demonstrate the BVP-constrained GPR framework on test problems in dimensions one and two. 

\subsection{One-dimensional Dirichlet Problem for the Laplacian}\label{sec:1d}

Suppose a function $u$ is constrained to satisfy the one-dimensional Dirichlet problem on $\Omega = [0,1]$,
\begin{equation}\label{eq:1d_problem}
\left\{
\begin{aligned}
-\frac{d^2 u}{dx^2} &= f(x), \quad x \in (0,1) \\
u(0) &= u(1) = 0,
\end{aligned}
\right.
\end{equation}
where only scattered noisy observations of $u$ and/or $f$ are available.
The corresponding eigenvalue problem is 
\begin{equation}
\left\{
\begin{aligned}
-\frac{d^2 \phi_n}{dx^2} &= \lambda_n \phi_n, \quad x \in (0,1) \\
\phi_n(0) &= \phi_n(1) = 0,
\end{aligned}
\right.
\end{equation}
which has the solution
\begin{equation}
\phi_n = \sqrt{2}\sin(\sqrt{\lambda_n}x), \quad \lambda_n = n^2 \pi^2
\end{equation}
for positive integers $n$.

We generated synthetic data points from $f(x)= x$ with solution $u(x) = -\frac{1}{6}(x^3 -x)$ and added normally distributed white noise. $M = 8$ eigenfunctions were used for the spectral expansion. To optimize the hyperparameters, the negative log-marginal-likelihood was minimized 1000 times using the L-BFGS-B method from random initial parameters. For each minimization, the initial $s^2$ was drawn from the exponential distribution with scale $1$, the initial $\ell$ was drawn from the uniform distribution on $[0,0.5]$, and the initial $\sigma$ was drawn from the uniform distribution on $[0,0.3]$. Bounds for the parameters were $[\text{1e-4, 1e4}]$ during the optimization. The hyperparameters corresponding to the least value of the negative log-marginal-likelihood minimum were selected.  

\begin{figure}
\centering
\includegraphics[width=.49\linewidth]{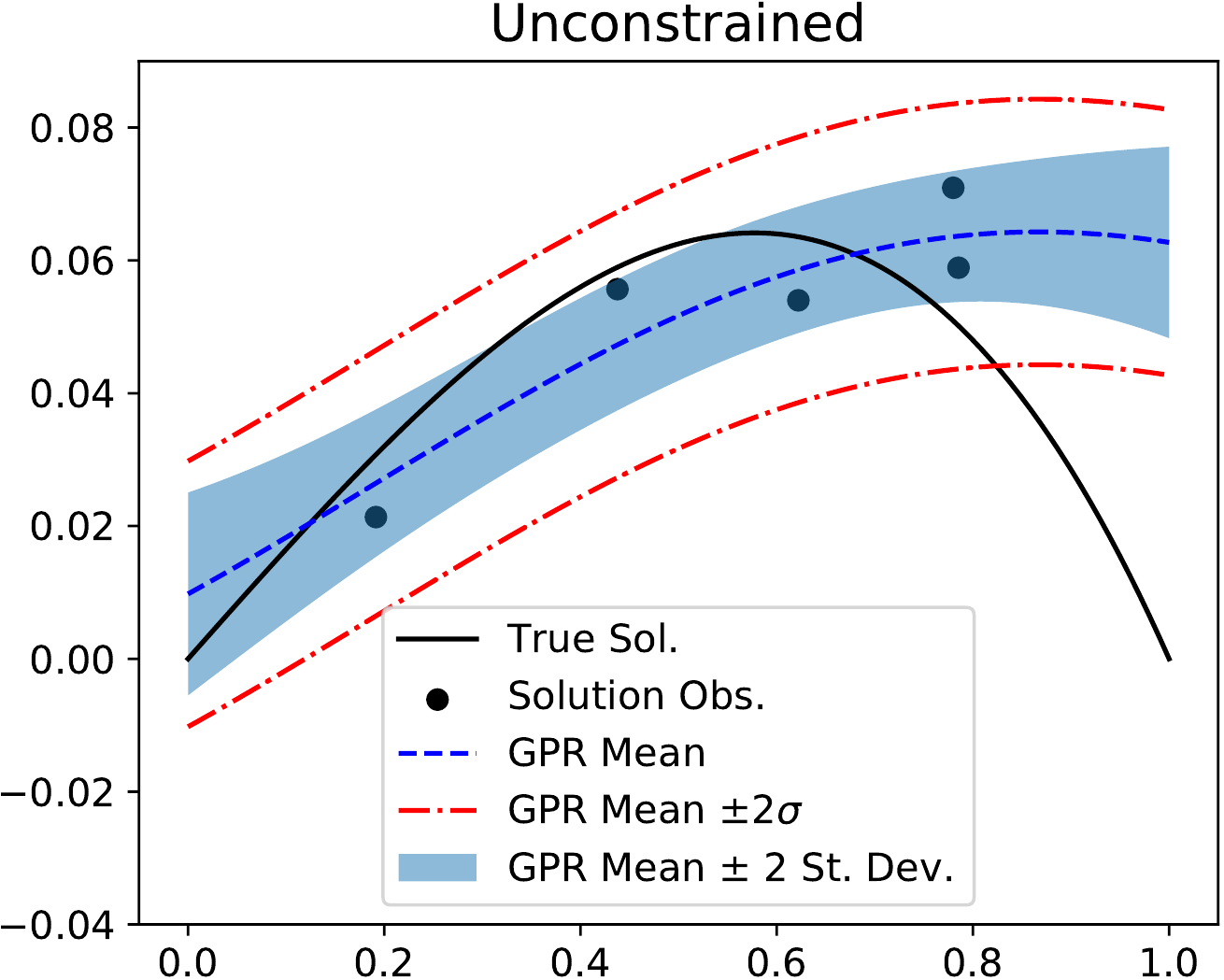}
\includegraphics[width=.49\linewidth]{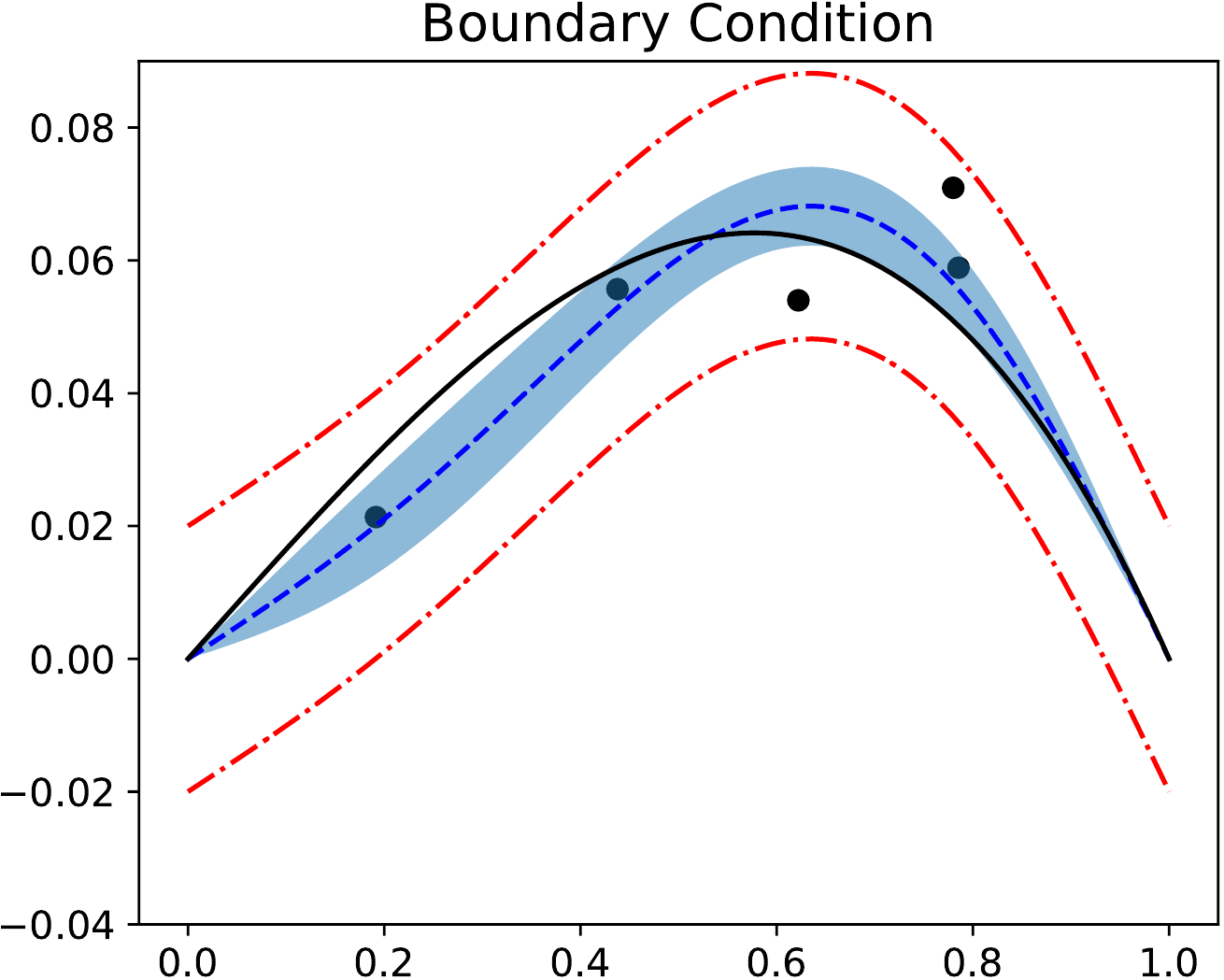}\\
\vspace{4ex}
\includegraphics[width=.49\linewidth]{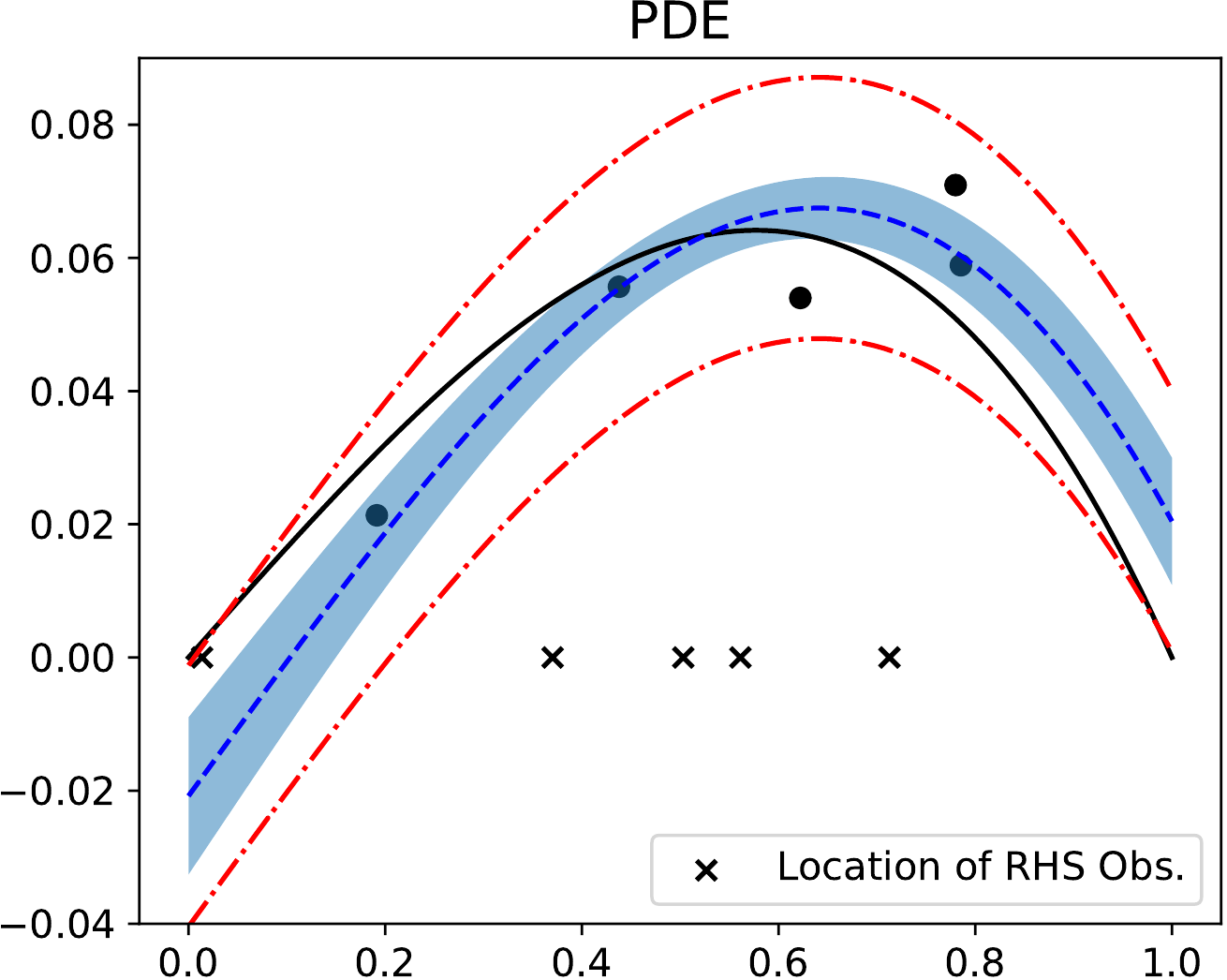}
\includegraphics[width=.49\linewidth]{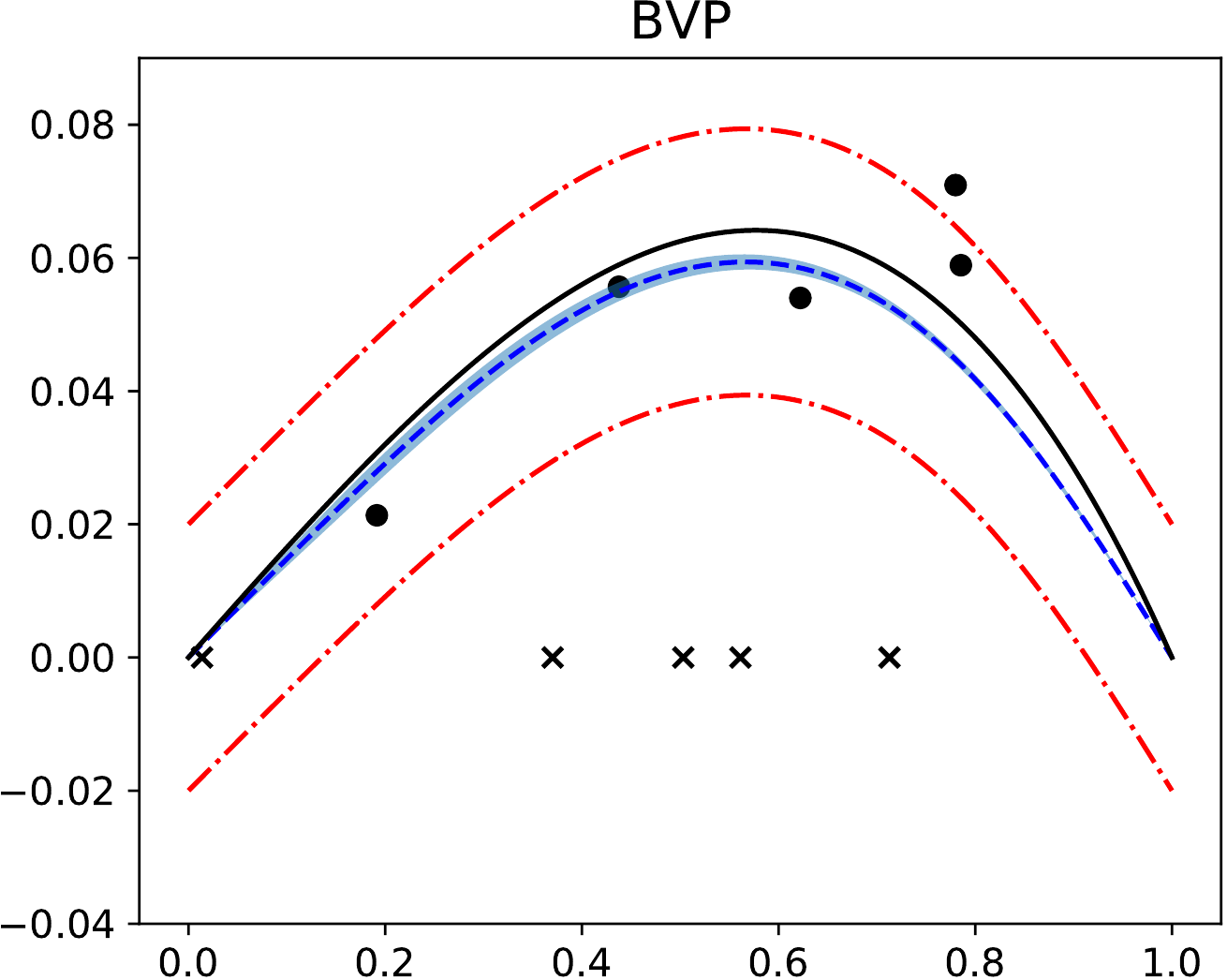}
\caption{Comparison of unconstrained GPR with three types of constrained GPR. 
\emph{Top Left:} Unconstrained GPR using a standard squared-exponential kernel; relative $\ell^2$ error of 42.5\%. \emph{Top Right:} BC-constrained GPR using the spectral expansion kernel; relative $\ell^2$ error of 14.6\%. \emph{Bottom Left:} PDE-constrained GPR using a squared-exponential kernel; relative $\ell^2$ error of 25.9\%. \emph{Bottom Right:} BVP-constrained GPR; relative $\ell^2$ error of 9.3\%. The relative errors are between the posterior mean of the GPR (dashed blue curve) and the exact solution $u$ (solid black curve). Each example uses the same 5 observations (black dots) of the function $u$ at randomly sampled points in $[0,1]$, obtained by sampling $u$ and adding white noise with $\sigma = 0.01$. 
The PDE and BVP constrained problems also involve 5 observations of $f$, the right hand side of the PDE $-u'' = f$, at randomly sampled points indicated by black ``x'' marks. The shaded blue region indicates twice the posterior standard deviation around the mean. The dashed red lines show twice the learned noise parameter around the posterior mean.
}\label{fig:dim1_comparison}
\end{figure}

We performed two trials. The first is a comparison of four methods when observations of both $u$ and $f$ are available: unconstrained GPR using the squared-exponential kernel function as described in Section \ref{sec:gpr}, boundary condition constrained GPR as described in Section \ref{sec:bc_constrained_gpr}, PDE constrained GPR as described in Section \ref{sec:pde_constrained_gpr}, and finally the BVP constrained GPR developed in Section \ref{sec:bvp_constrained_gpr} which combines both boundary condition and PDE constraints. The unconstrained GPR was performed using the \texttt{scikit-learn} package. We provide $5$ scattered observations of $u$ at $x = 0.19, 0.44, 0.62, 0.78, 0.79$, and apply a white noise with standard deviation $\sigma = 0.01$. For PDE and BVP constrained GPR, we also provide $5$ scattered observations of $f$ at $x = 0.01, 0.37, 0.50, 0.56, 0.71$ polluted by white noise with the same standard deviation $\sigma = 0.01$. Figure \ref{fig:dim1_comparison} provides a comparison of the resulting regression of $u$. Qualitatively, when boundary condition constraints are provided, the mean prediction satisfies the boundary condition while the variance is zero at the boundaries. The combined BVP constrained GPR exhibits the smallest uncertainty and the smallest error between the true solution $u$ and the mean prediction $u^{*}$, measured in the relative $\ell^2$ error over $100$ uniformly spaced test points on $[0,1]$,
\begin{equation}
\text{error} = \frac{\| u^* - u \|_{\ell^2}}{\|u\|_{\ell^2}}.
\end{equation}

\begin{figure}
\centering
\includegraphics[width=.49\linewidth]{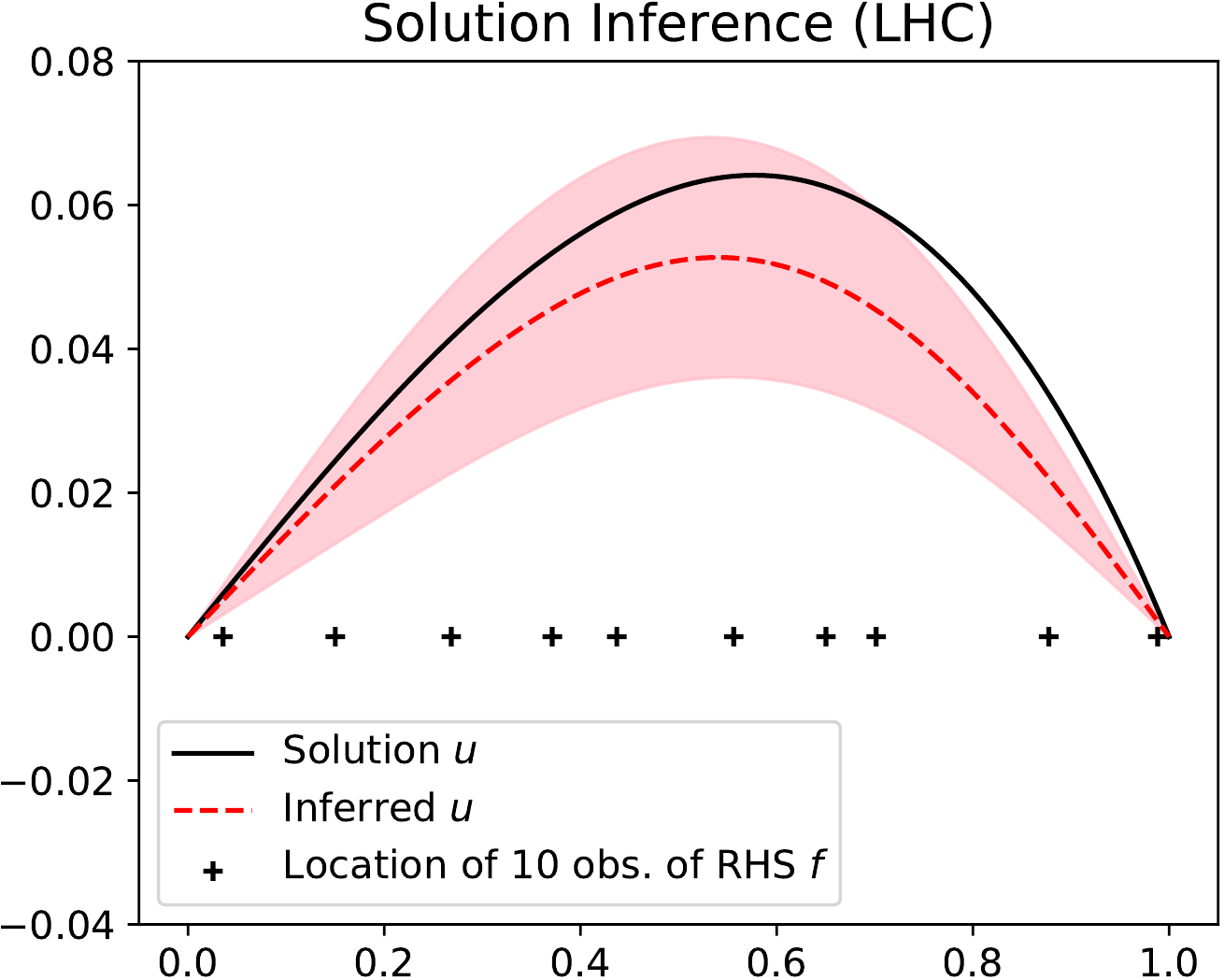}
\includegraphics[width=.49\linewidth]{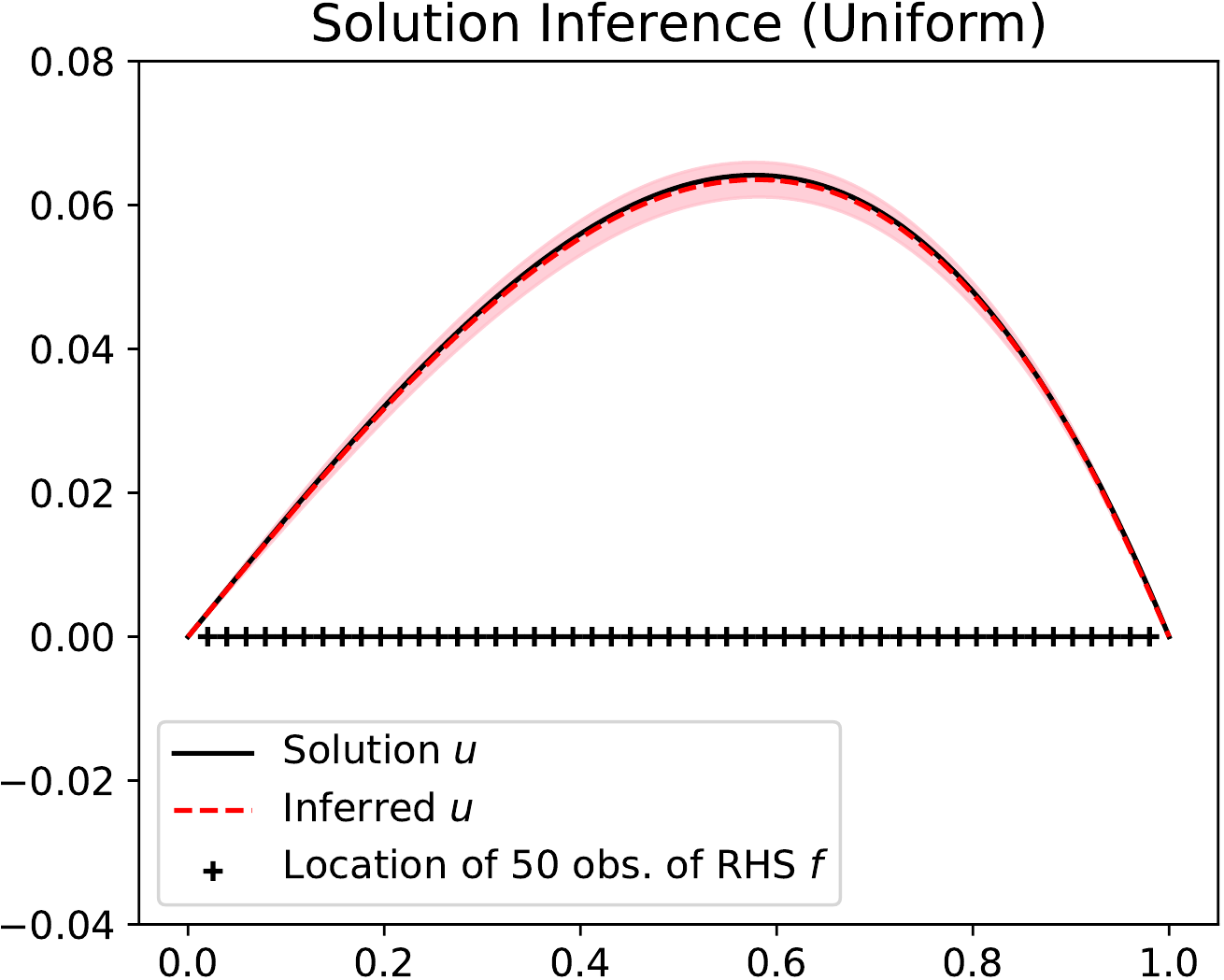}
\caption{Illustration of inferring the solution $u$ to the BVP \eqref{eq:1d_problem} with noisy observations of $f$ at the indicated locations, using BVP constrained GPR. The locations are sampled by maximin Latin hypercube sampling (LHC, \emph{left}) or lie on a uniform grid (\emph{right}). As the number of observations increases, the trend is for the mean prediction $u^*$ (dashed red line) to be closer to the true solution $u$ (solid black line), with the posterior variance (shaded red region) growing smaller.}
\label{fig:dim1_inference_illustration}
\end{figure}

\begin{figure}
\centering
\includegraphics[width=.49\linewidth]{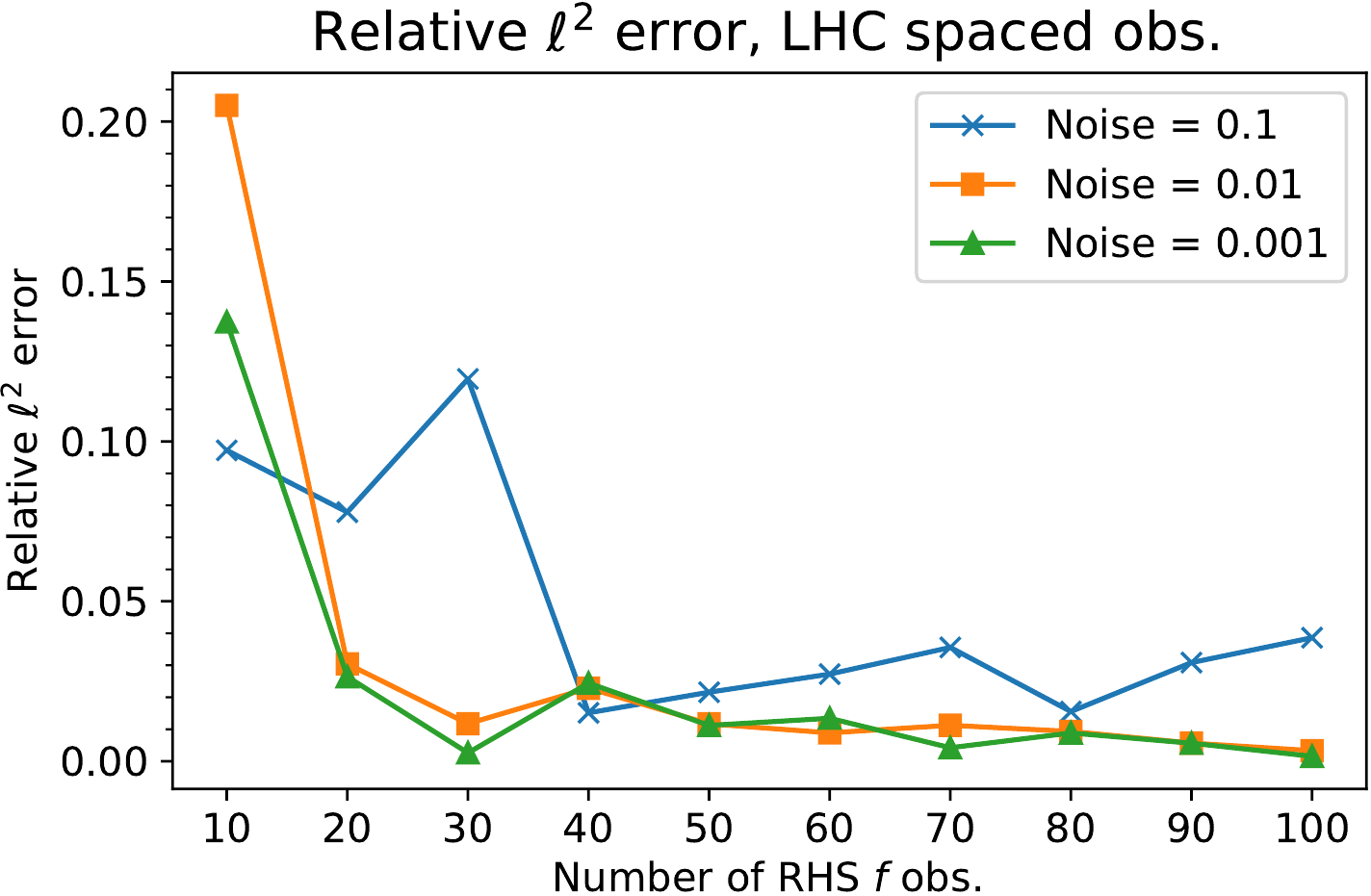}
\includegraphics[width=.49\linewidth]{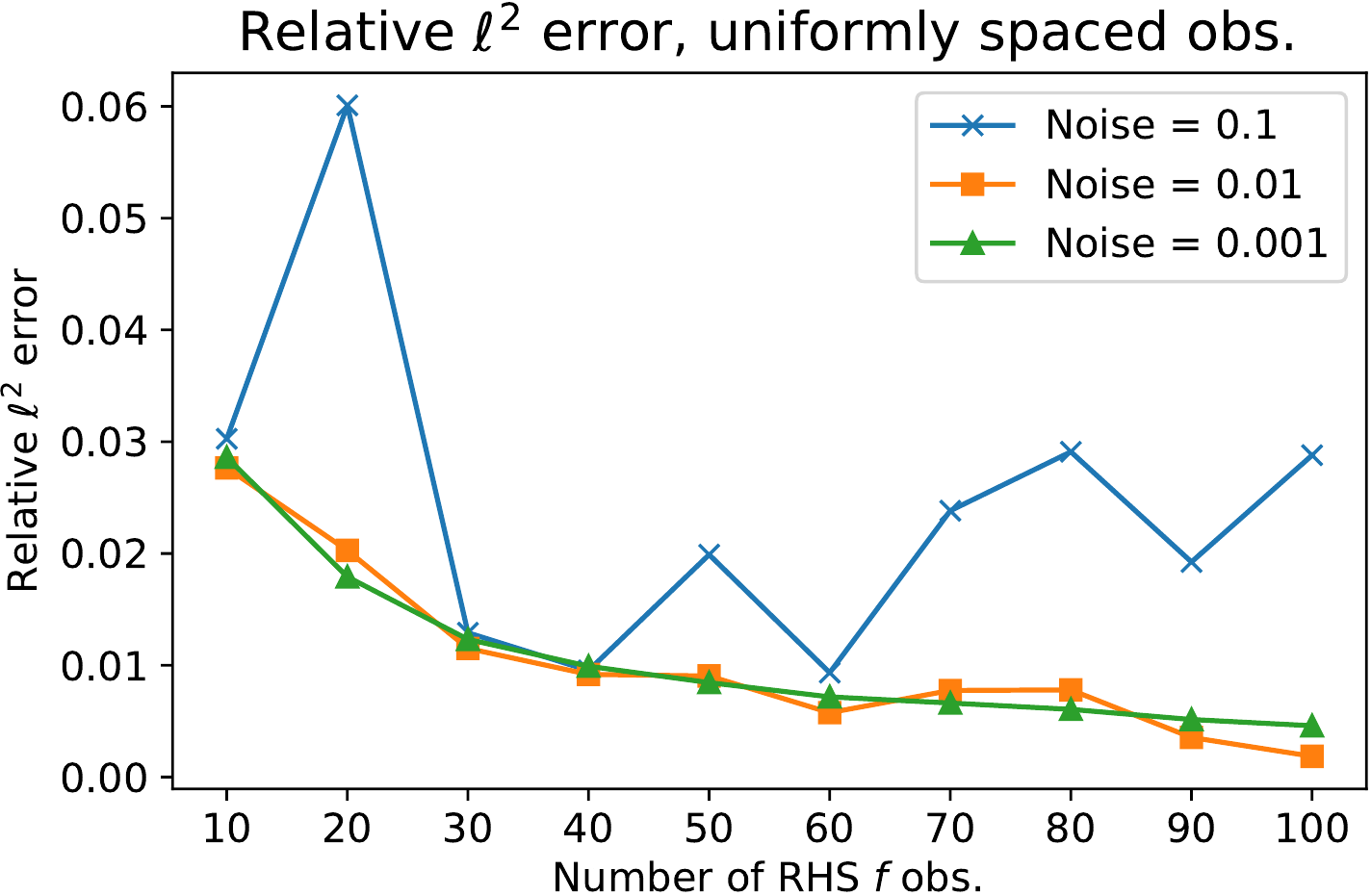}
\caption{Plot of the error between the posterior mean prediction $u^*$ and the true solution $u$, measured in the relative $\ell^2$ norm over 100 uniformly spaced test points in $[0,1]$. For the relatively large value of white noise standard deviation $\sigma = 0.1$ (applied to observations of $f$), the trend is less consistent, but for $\sigma = 0.01$ and $\sigma = 0.001$ the error trends more consistently and saturates around $1\%$ for both observations at LHC sampled locations and on the uniform grid.}
\label{fig:dim1_inference_convergence}
\end{figure}

In the second trial, we consider the case where only observations of the source term $f$ are available, and inference is required for $u$. In this case, only the PDE constrained GPR and the BVP constrained GPR may be useful. We generate $n_f$ observations of $f$ by sampling locations in $[0,1]$ either randomly using Latin hypercube (LHC) maximin sampling via the \texttt{pyDOE} package, or using a regular (uniform) grid on $[0+1/n_f,1-1/n_f]$. We apply white noise with standard deviation $\sigma = 0.001, 0.01$, or $0.1$ as the case may be. We found that the BVP constrained GPR for $u$ can be unstable for $n_f < M$; however, in this example, it provides accurate inference of $u$ using 10 or more observations. We illustrate this in Figure \ref{fig:dim1_inference_illustration}. We also study the relative $\ell^2$ error between the inferred solution $u^*$ and the true solution $u$ as $n_f$ increases, for all three values of white noise standard deviation used to generate the data. Figure \ref{fig:dim1_inference_convergence} illustrates that for $\sigma = 0.01$ and $\sigma = 0.1$, the error behaves consistently as $n_f$ increases, exhibiting a decreasing trend and saturating at about $1\%$ relative error. 

\begin{figure}
\centering
\includegraphics[height=2.8in]{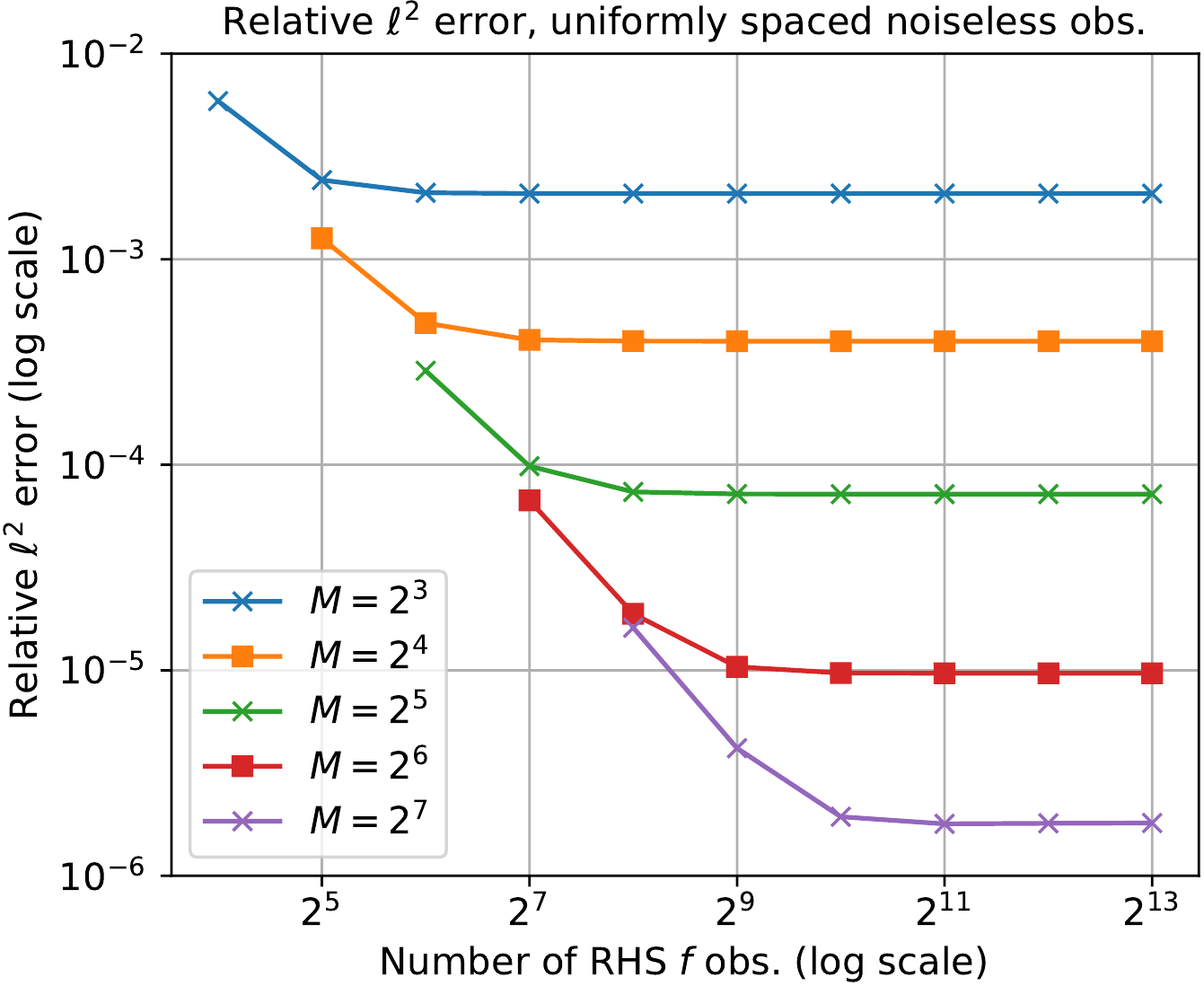}
\includegraphics[height=2.8in]{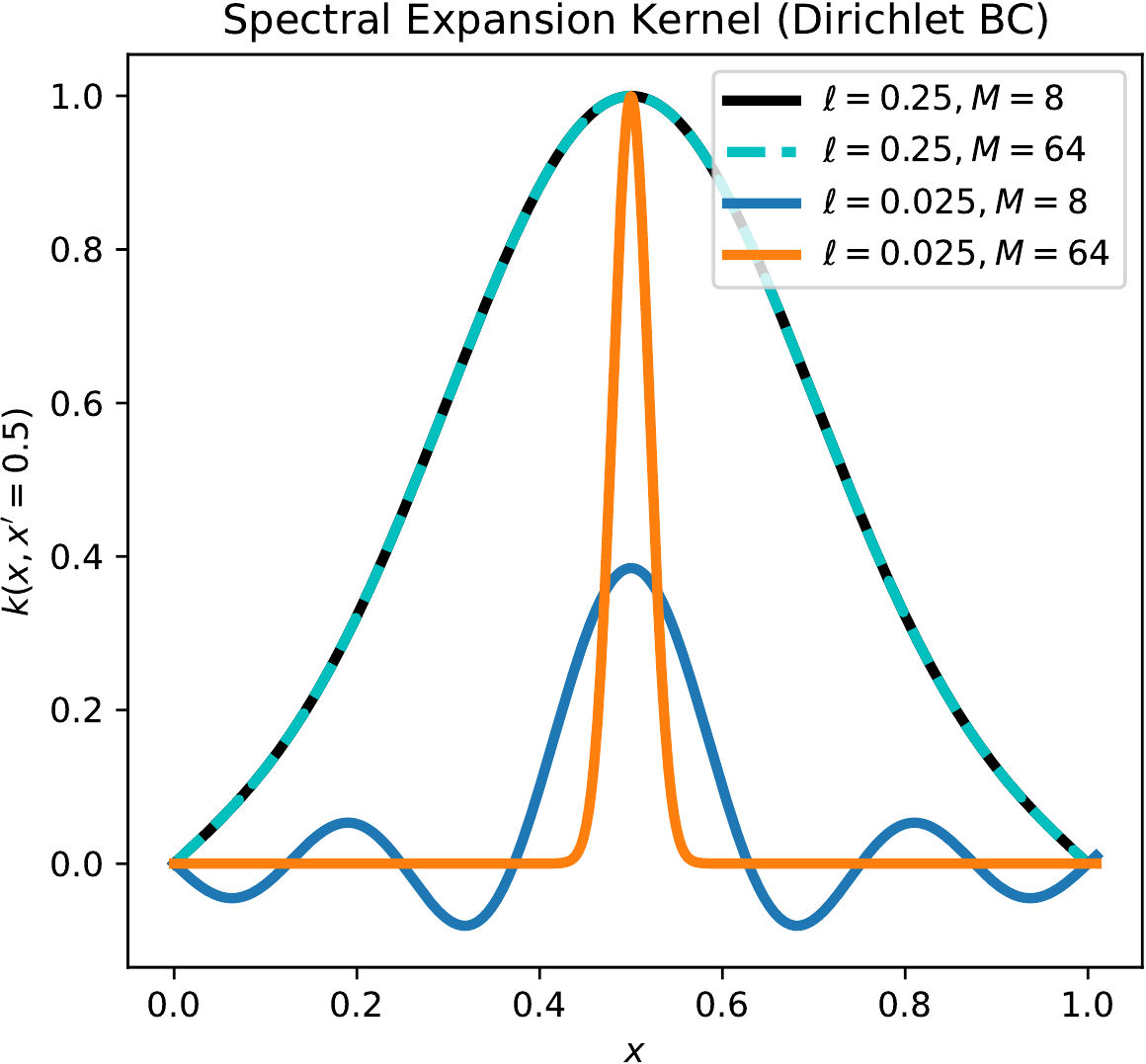}
\caption{\emph{Left:} Convergence in log-log scale of the error between the posterior mean prediction $u^*$ and the true solution $u$, trained with noiseless observations, measured in the relative $\ell^2$ norm over 100 uniformly spaced test points in $[0,1]$. The noise/likelihood hyperparameter $\sigma$ is fixed to $10^{-17}$. For fixed number $M$ of eigenfunctions defining the covariance kernel, the error decreases with the number $n_f$ of observations. As $M$ increases, the error decreases. \emph{Right:} Plotting the spectral expansion covariance kernel $k(x,x'=0.5)$ for various $M$ reveals that artifacts are present when the correlation length hyperparameter $\ell$ (width of the parent squared exponential kernel) is small, and increasing $M$ reduces these artifacts.}\label{fig:dim1_noiseless_convergence}
\end{figure}

The saturation of error for the case of noisy observations raises the question of whether the BVP-GP method can yield convergent error in solving the BVP \eqref{eq:1d_problem} given noiseless observations of the source $f$. We therefore applied the BVP-GP method with increasing number $n_f$ of noiseless observations, with the noise/likelihood hyperparameter fixed to be $10^{-17}$ (removing the $\sigma^2 I$ term from this experiment) rather than being trainable. The observations are located at a hierarchical sequence of uniform grids with $n_f = 2^p$ points, with $p \le 13$, and due to noiseless data, we require $n_f \ge M$. Our study in Figure \ref{fig:dim1_noiseless_convergence} reveals that for fixed $M$, as $n_f$ increases, the error saturates, but as $M$ increases, the error decreases for all $n_f$, and the limiting error decreases consistently as well. Given the representation \eqref{eq:representer} of the posterior mean $u^*$, to obtain a more accurate regression of $u$ as the density of observations increases, smaller correlation lengths $\ell$ in the covariance kernels \eqref{eq:expansion_kernel} should be utilized \citep{rasmussen,gyorfi2006distribution,stein2012interpolation}. This parameter corresponds to the width $\ell$ of the squared-exponential kernel that gives rise to the spectral covariance kernels via \eqref{eq:spectral_density}. However, we show in Figure \ref{fig:dim1_noiseless_convergence} that for small $\ell$, the spectral covariance kernels may exhibit oscillations away from the peak that pollute the global approximation quality, causing the error saturation. As $M$ increases, these artifacts are reduced, which explains why the error saturates at a smaller level for larger $M$ in Figure \ref{fig:dim1_noiseless_convergence}, and converges as $M,n_f \rightarrow \infty$. Of course, when an abundance of noiseless observations of $f$ are available, a more traditional numerical method for solving the BVP, admitting a more straightforward analysis of accuracy and stability, may be preferable. The reduced-rank property of BVP-GPR was apparent in performing this study; even with $8192$ observations, the entire training and inference process took around one minute using an Intel(R) Core(TM) i7-8650U CPU @ 1.90GHz.

\begin{figure}
\centering
\includegraphics[width=.31\linewidth]{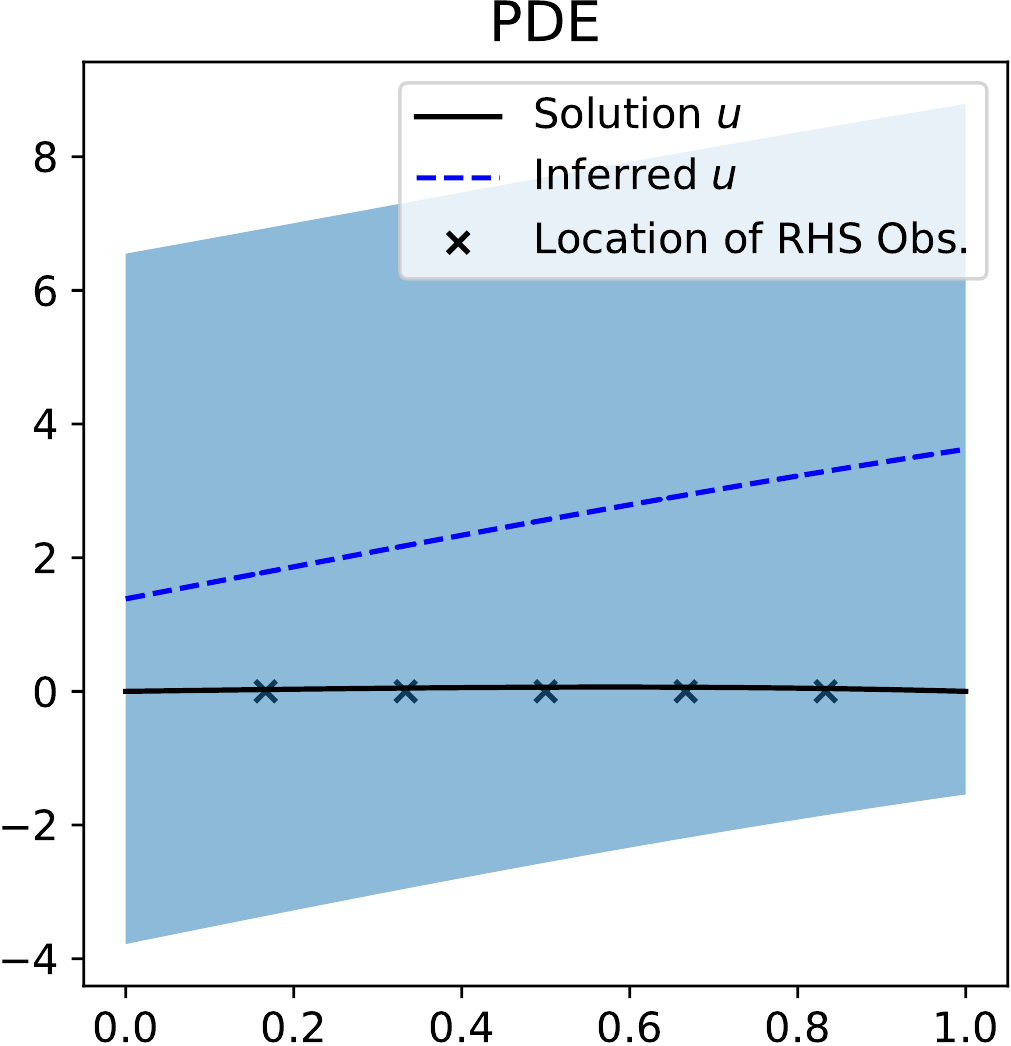}
\includegraphics[width=.33\linewidth]{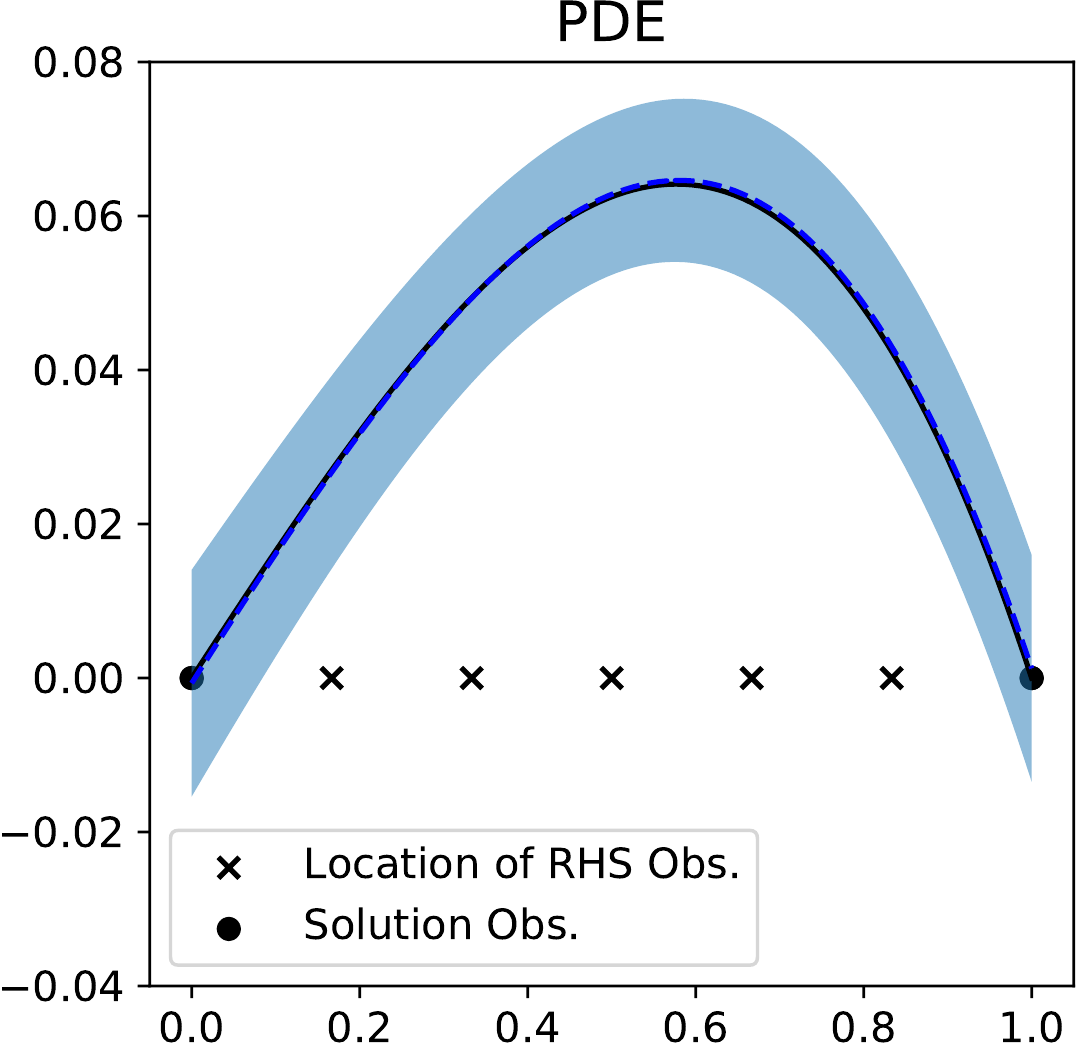}
\includegraphics[width=.33\linewidth]{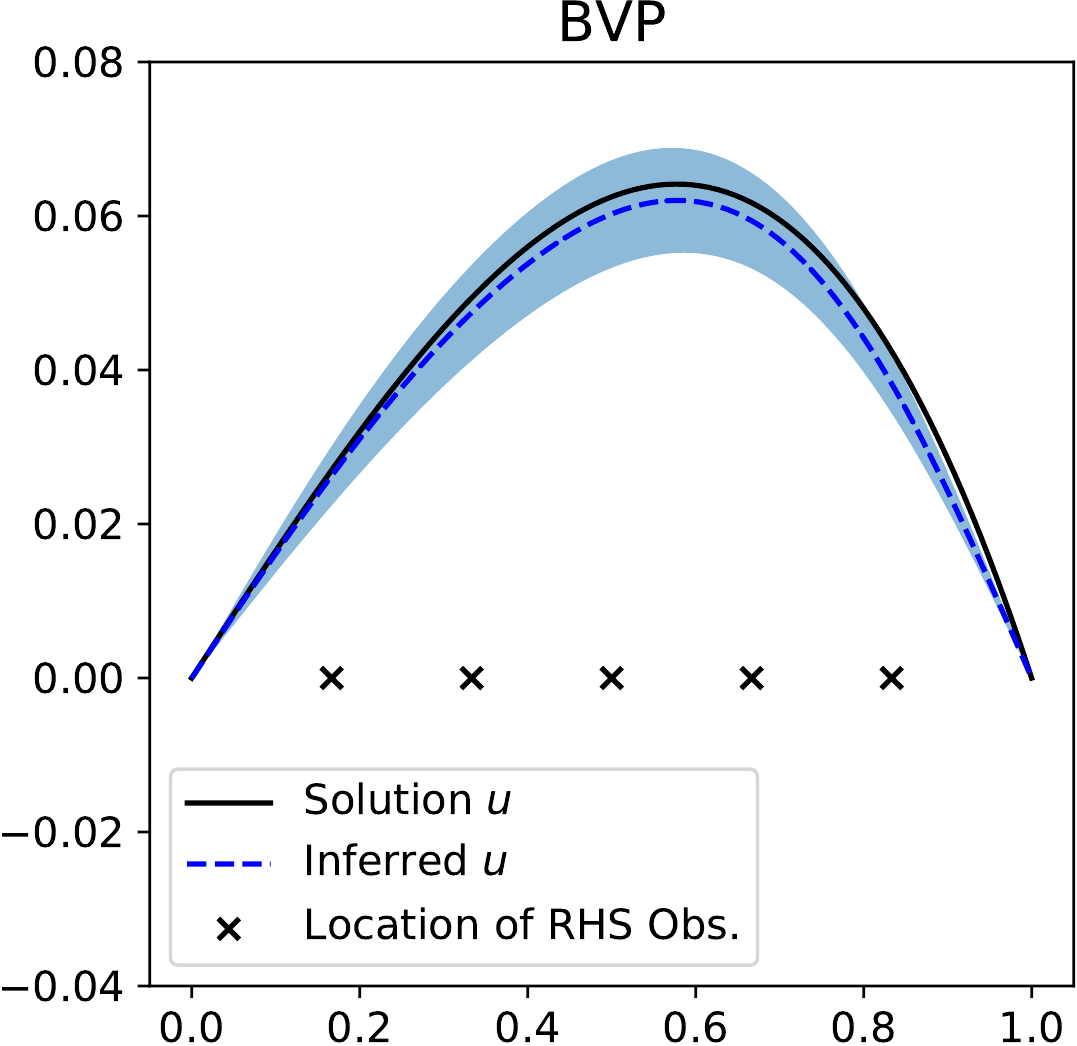}
\caption{Effect of enforcing the boundary conditions when inferring the solution $u$ from five scattered observations of $f$. When using the PDE-GP method \emph{(left)}, inference fails without observations of $u$, as even with complete knowledge of $f$, $u$ is only determined up to an arbitrary linear function. When the boundary conditions are treated in the one-dimensional PDE-GP method as point observations of $u$ \emph{(center)}, accurate inference is possible although uncertainty is nonzero in contrast to the BVP-GP method. In the BVP-GP method \emph{(right)}, the boundary conditions are enforced with certainty via the covariance kernel, not as discrete observations, which is advantageous in higher dimensions.}
\label{fig:dim1_illposedness}
\end{figure}

In contrast to the BVP-GP method, solution inference using only the PDE constraints (PDE-GP) without observations of $u$ fails regardless of $n_f$, yielding an inferred solution $u^*$ that differs from $u$ by orders of magnitude. This is illustrated in Figure \ref{fig:dim1_illposedness}. This can be understood by considering the ideal case of infinitely many noiseless observations of $f$, in the fixed-domain asymptotic regime \citep{rasmussen,gyorfi2006distribution,stein2012interpolation}. Although this would allow for recovering $f$ to any accuracy, the solution $u$ would only be determined up to an arbitrary linear function. When the boundary condition is utilized by adding observations of $u$ at the two boundary points as in Figure \ref{fig:dim1_illposedness}, the PDE-GP method then yields an accurate prediction; however, variance at the boundaries is nonzero, unlike for the BVP-GP method. Moreover, in higher dimensions, enforcing the boundary condition through scattered point observations in this way would rapidly increase the cost and ill-conditioning of the training and inference steps of GPR. 

\subsection{Two-dimensional Helmholtz Equation}\label{sec:2d}

Now we consider the solution of the Helmholtz equation
\begin{equation}\label{eq:helmholtz_equation}
-\nabla^2 u(x,y) + k^2 u(x,y) = f(x,y), \quad (x,y) \in (0,1) \times (0,1), \\
\end{equation}
with $k=3$ subject to the mixed boundary condition,
\begin{equation}
\left\{
\begin{aligned}
\frac{\partial u}{\partial x}(x=0,y) &= 0, \\
\frac{\partial u}{\partial y}(x,y=0) &= 0, \\
u(x=1,y) &= 0, \\
u(x,y=1) &= 0.
\end{aligned}
\right.
\end{equation}
We generate observations by sampling the solution $u$ and source term $f$,
\begin{align}
\begin{split}
&u(x,y) = (1-x^2)(1-y^2) + \cos\left(\frac{\pi x}{2}\right)\big(\exp(-y)+y-(1+\exp(-1))\big) \\
&f(x,y) = 2(1-x^2)+2(1-y^2)+\left(\frac{\pi}{2}\right)^2 \cos\left(\frac{\pi x}{2}\right) \big( \exp(-y) + y - 1+\exp(-1) \big) \\
&\phantom{=}-
\cos\left(\frac{\pi x}{2} \right)\exp(-y) + k^2 \big[ (1-x^2)(1-y^2)+\cos\left(\frac{\pi x}{2}\right) \big( \exp(-y)+y-1+\exp(-1) \big) \big]
\end{split}
\end{align}
at random locations obtained by maximin Latin hypercube sampling on $[0,1] \times [0,1]$ and adding white noise with standard deviation $\sigma = 0.01$.

The Helmholtz operator is positive definite with corresponding multi-indexed eigenvalues
\begin{equation}
\lambda_{mn} = \mu_m^2 + \nu_n^2 + k^2
\end{equation}
\begin{equation}
\mu_m = \frac{(2m+1)\pi}{2}
\end{equation}
\begin{equation}
\nu_n = \frac{(2n+1)\pi}{2}
\end{equation}
and eigenfunctions
\begin{equation}
\Phi_{mn}(x,y) = g_m(x)h_n(y)
\end{equation}
\begin{equation}
g_m(x) = \cos(\mu_m x)
\end{equation}
\begin{equation}
h_n(y) = \cos(\nu_n y)
\end{equation}
Because of the multiple dimensions, the number of eigenfunctions to track increases quadratically with the index, so with $M_1$ eigenfunctions per dimension there are $M = (M_1)^2$ degrees of freedom to model, and GP inference scales as $M^3 = (M_1)^6$. Nevertheless, if $M$ is much less than the number of data points, the cost of inference in the spectral approach is still low.

\begin{figure}
\centering
\includegraphics[width=.43\linewidth]{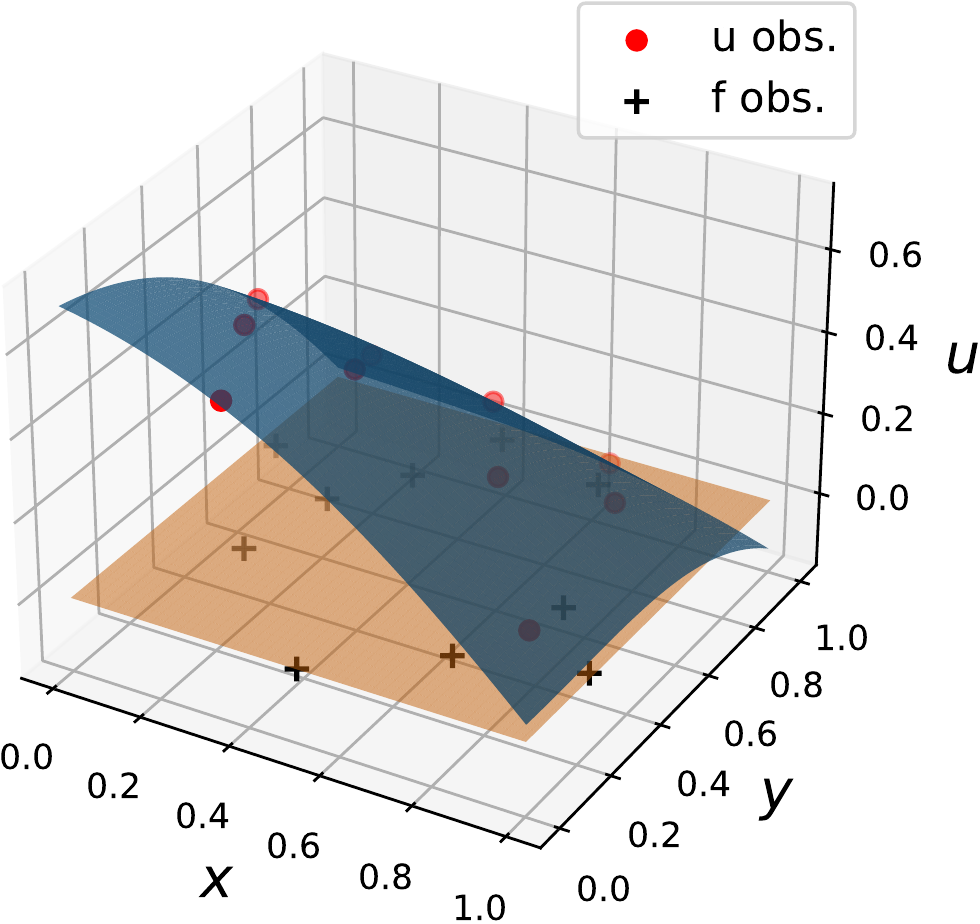}
\hspace{0.03\linewidth}
\includegraphics[width=.43\linewidth]{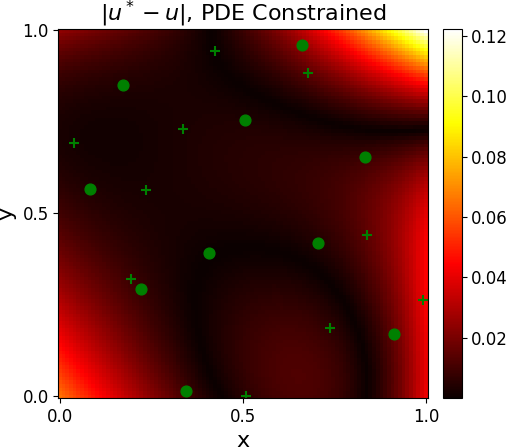}\\
\vspace{4ex}
\includegraphics[width=.43\linewidth]{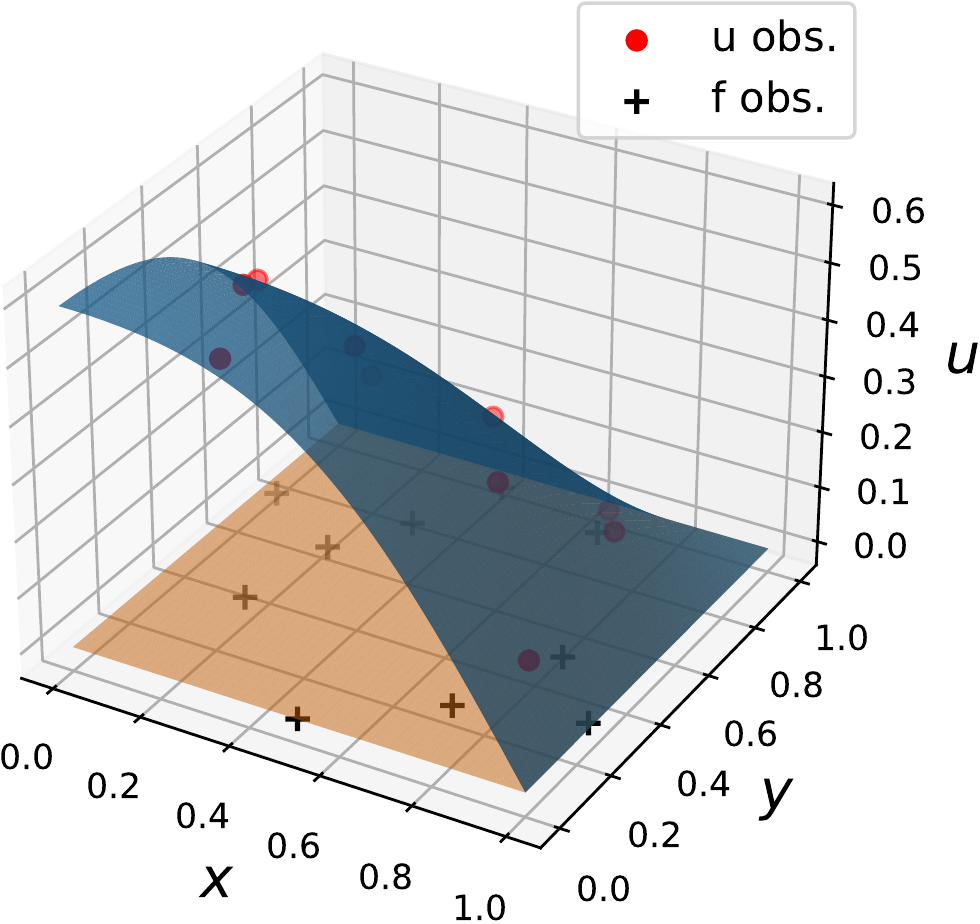}
\hspace{0.03\linewidth}
\includegraphics[width=.43\linewidth]{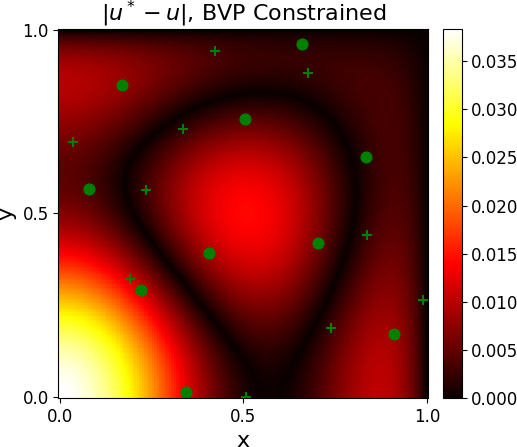}
\caption{Comparison of PDE constrained GPR (\emph{top}) and BVP constrained GPR (\emph{bottom}). The left column shows observations of $u$ (red dots) and locations of the observations of the source $f$ (black crosses) and the resulting mean prediction surface $u^*$ (blue). The $xy$-plane is plotted in orange as a reference for observing the boundary behavior of $u^*$. The right column plots the absolute error between the mean prediction $u^*$ and the true solution $u$. The BVP constrained GPR demonstrates lower local error and enforcement of the boundary conditions compared to the PDE constrained GPR. It also obtains a lower relative $\ell^2$ error over the uniform $100 \times 100$ test grid: 2.88\% vs 5.25\%.}
\label{fig:dim2_surfs_error}
\end{figure}

\begin{figure}
\centering
\includegraphics[width=.43\linewidth]{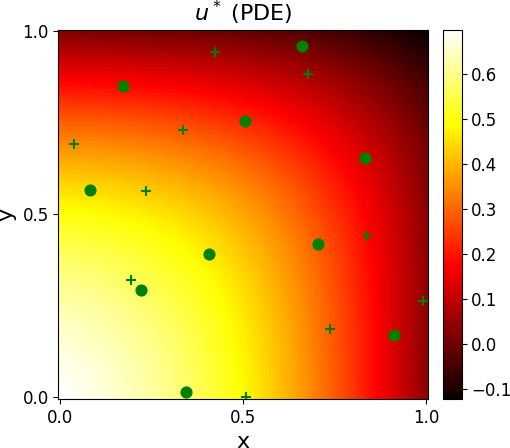}
\hspace{0.03\linewidth}
\includegraphics[width=.43\linewidth]{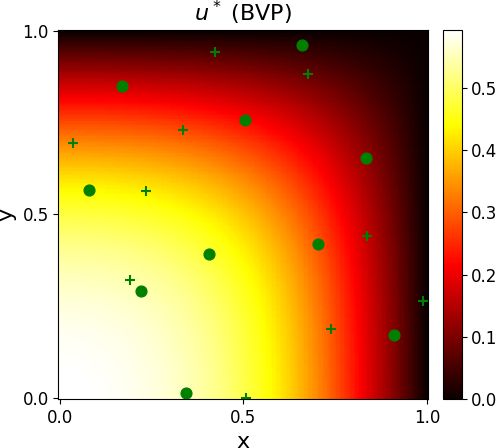}\\
\vspace{4ex}
\includegraphics[width=.43\linewidth]{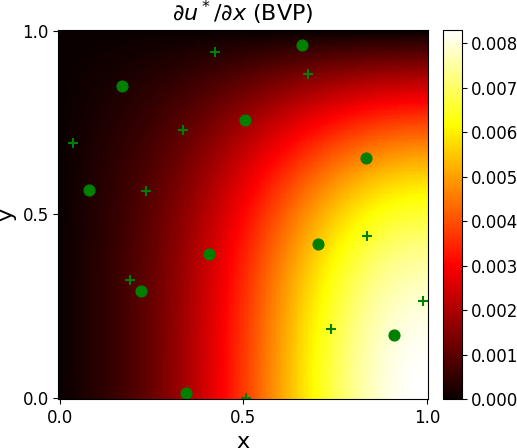}
\hspace{0.03\linewidth}
\includegraphics[width=.43\linewidth]{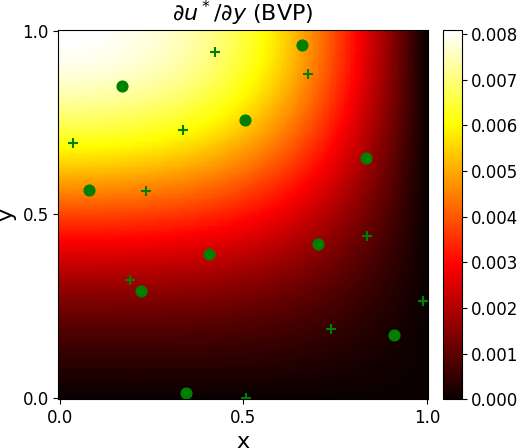}
\caption{Illustration of the boundary behavior of the GPR prediction $u^*$. In the top row, we compare the boundary values of the GPR prediction $u^*$ for the PDE constrained method (\emph{left}) and the BVP constrained method (\emph{right}). Unlike the PDE constrained $u^*$, the BVP constrained $u^*$ goes to zero at the two sides $(x = 1,y)$ and $(x,y=1)$ of the domain $[0,1] \times [0,1]$.
In the bottom row, we verify that that the BVP constrained $u^*$ satisfies the Neumann condition at the other sides. The left panel shows $\partial u^*/\partial x$, and demonstrates that 
$\partial u^*/\partial n = 0$ over $(x=0,y)$, while the right panel shows $\partial u^*/\partial y$ and demonstrates that $\partial u^*/\partial n = 0$ over $(x,y=0)$.
}
\label{fig:dim2_bcs}
\end{figure}

We demonstrate the BVP constrained framework and compare with the PDE constrained framework below. We sample 10 observations each for of $u$ and $f$. We use a kernel with $M_1=3$ eigenfunctions per dimension. We apply maximum likelihood estimation via the L-BFGS-B algorithm with 100 random initial hyperparameters as in Section \ref{sec:1d}. The same bounds of $[\text{1e-4,1e4}]$ are enforced during the training, and we employ the optimal final hyperparameters for inference. Our results in Figure \ref{fig:dim2_surfs_error} demonstrate a significant benefit in fidelity using the BVP constrained GPR compared to PDE constrained GPR with the same dataset. Figure \ref{fig:dim2_bcs} provides a closer look and validation of the global enforcement of the boundary conditions of the GPR prediction.  

\begin{figure}
\centering
\includegraphics[width=.49\linewidth]{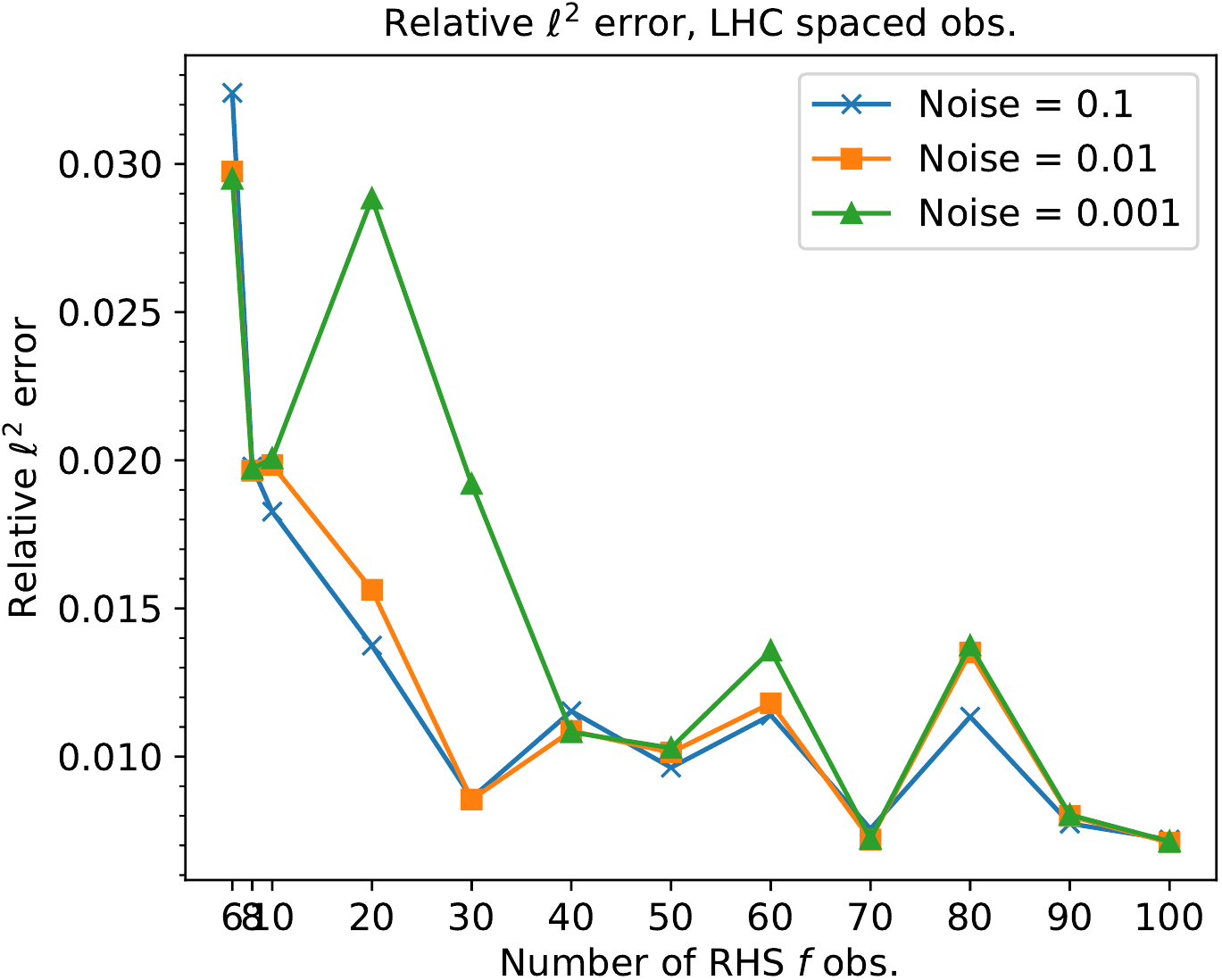}
\includegraphics[width=.49\linewidth]{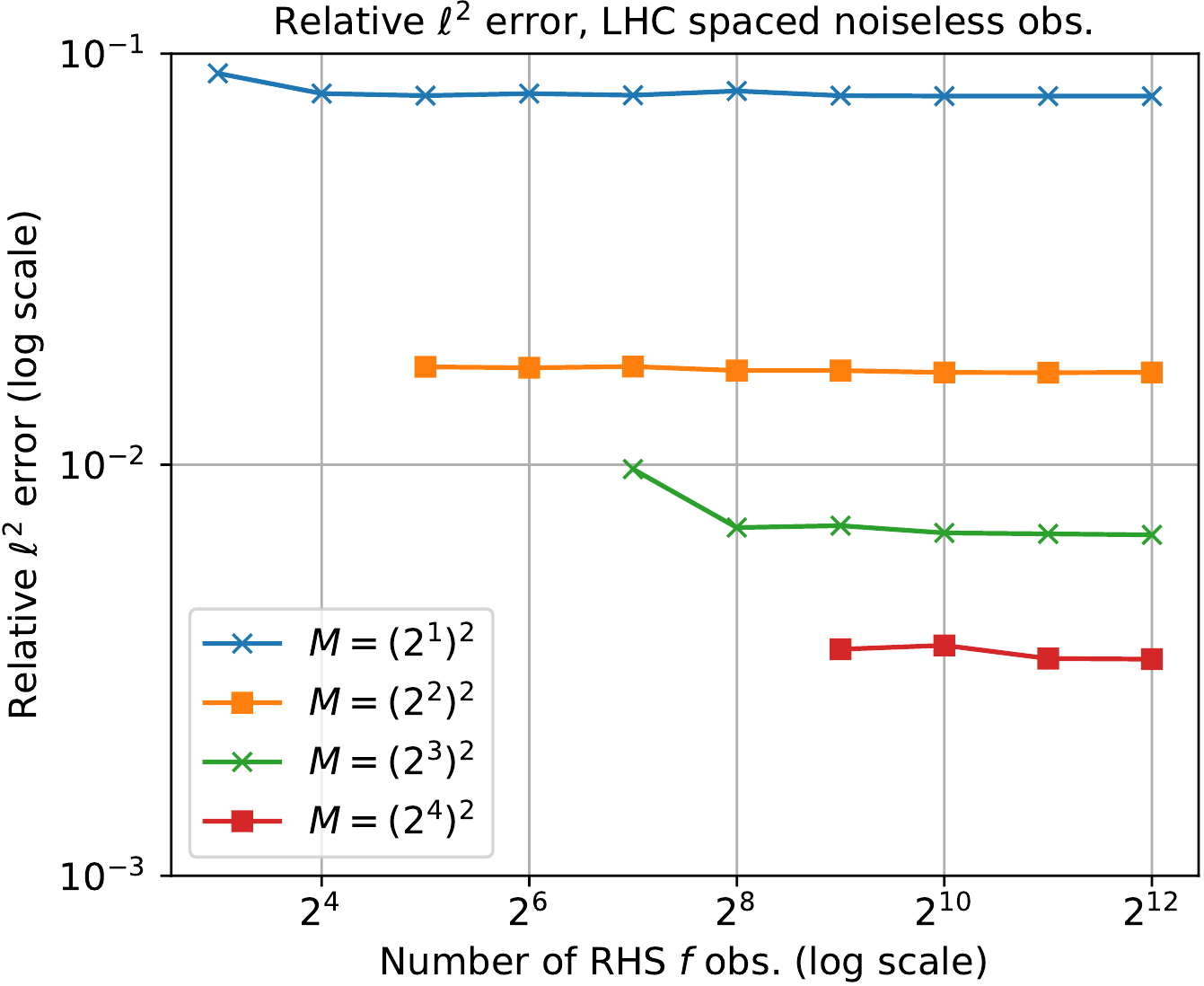}
\caption{\emph{Left:} Plot of the error between the posterior mean prediction $u^*$ and the true solution $u$ of the Helmholtz equation with mixed boundary condition, measured in the relative $\ell^2$ norm over a uniform $100 \times 100$ uniform test grid. The $n_f$ observations of the source term $f$ are sampled from maximin Latin hypercube sampling; no observations of $u$ are given. \emph{Right:} Convergence, in log-log scale, of the same quantity when trained using noiseless observations of $f$ and noise/likelihood hyperparameter $\sigma$ fixed as $10^{-17}$ for increasing total number $M$ of eigenfunctions defining the covariance kernel.}
\label{fig:dim2_error}
\end{figure}

Next, we study the use of BVP constrained GPR to infer the solution $u$ to the Helmholtz problem \eqref{eq:helmholtz_equation} when observations of $u$ are not available. We consider 
\begin{equation}
n_f = 6,8,10,20,30,40,50,60,70,80,90,100
\end{equation}
observations of $f$ located at random locations obtained via the maximin Latin hypercube sampling. We pollute these observations with white noise of three standard deviations $\sigma = 0.1, 0.01, 0.001$ as the case may be. We apply the same procedures for training and inference as described above, and plot the relative $\ell^2$ error between the GPR prediction $u^*$ and the true solution $u$ on a uniform $100 \times 100$ test grid in Figure \ref{fig:dim2_error}. We see again that the error saturates on the order of $1\%$, in this case around $n_f \sim 40$. In the same figure, we also show a convergence study for the case of noiseless observations of $f$ with increasing $M = (M_1)^2$ total eigenfunctions, training with fixed noise/likelihood parameter $\sigma = 10^{-17}$ and requiring $n_f \ge M$. For this case, we sample observations at a non-hierarchical sequences of Latin hypercube grids with $2^p$ elements with $p \le 12$. We observe decreasing error as $M,n_f$ increase, as for the one-dimensional study, although more slowly in comparison.  

\section{Conclusion}\label{sec:conclusion}
In this work we have developed a framework that combines the use of spectral decomposition covariance kernels with differential equation constraints in a co-kriging setup to perform Gaussian process regression constrained by boundary value problems. This BVP-GP approach constructs a GP that intrinsically satisfies the boundary condition while utilizing knowledge of the governing equation and observations of the source/forcing term. We tested the approach on benchmark boundary value problems in one and two dimensions. Our work includes the novel application of Gaussian process regression to boundary value problems with Neumann boundary conditions and to the case of inferring the solution $u$ of a boundary value problem from knowledge of the boundary condition and scattered observations of the source term alone. The lower-dimensional representation inherent to the spectral covariance kernel yielded an efficient training and inference process, which allowed us to perform convergence studies of the error in inferring the solutions of boundary value problems with a number of observations (up to $8192$) that would have been prohibitive using a standard GPR approach, due to computational cost and ill-conditioning. Our studies showed that the BVP-GP method can be seamlessly used in a spectrum of applications from small datasets with high noise to large, noiseless datasets. The examples discussed lend themselves to analytical eigenfunction decompositions, but in more complex domains, numerically computed eigenfunctions may be substituted.

One limitation of this framework is that the spectral decomposition is limited to the setting where the parameters of the operator are known. In the Helmholtz equation, for example, the parameter $k^2$ was assumed to be known a priori. In cases where it is not known, it must be inferred, and this adds complexity to the evaluation of the eigenfunctions and eigenvalues, and to covariance operator as a whole. Although it is possible to incorporate the BVP operator parameters into the inference process, it is likely to increase the computational cost.
Another possible extension would be to incorporate time-dependence into the problem with temporally evolving spatial fields, as in time-dependent diffusion or wave propagation. This may involve a spectral decomposition in which the associated time-evolving magnitudes of the eigenfunctions are themselves governed by Gaussian processes as well, and their magnitudes must be inferred from the temporally correlated observations.

\section*{Acknowledgements}

This work was supported by the LDRD program at Sandia National Laboratories, and its support is gratefully acknowledged.
M. Gulian was also supported by the John von Neumann fellowship at Sandia National Laboratories, and by the U.S. Department of Energy, Office of Advanced Scientific Computing Research under the Collaboratory on Mathematics and Physics-Informed Learning Machines for Multiscale and Multiphysics Problems (PhILMs) project.
Sandia National Laboratories is a multimission laboratory managed and operated by National Technology and Engineering Solutions of Sandia, LLC., a wholly owned subsidiary of Honeywell International, Inc., for the U.S. Department of Energy's National Nuclear Security Administration under contract {DE-NA0003525}.
The views expressed in the article do not necessarily represent the views of the U.S. Department of Energy or the United States Government. SAND number: {SAND2020-14048 O}.

\bibliographystyle{elsarticle-num-names} 
\bibliography{gpldrd} 

\end{document}